\documentclass[lettersize,journal]{IEEEtran}
\AtBeginDocument{%
  }

\usepackage{amsmath, amssymb, amsfonts}
\usepackage{array}
\usepackage[caption=false,font=normalsize,labelfont=sf,textfont=sf]{subfig}
\usepackage{textcomp}
\usepackage{stfloats}
\usepackage{url}
\usepackage{verbatim}
\usepackage[nocompress]{cite}
\usepackage[T1]{fontenc}
\usepackage{booktabs}
\usepackage{xcolor}
\usepackage{algorithm}
\usepackage{algorithmicx}
\usepackage{algpseudocode}
\usepackage{graphicx}
\usepackage{lipsum}
\usepackage{enumerate}
\usepackage{enumitem}
\usepackage{hyperref}
\usepackage{multirow, makecell}
\usepackage{nccmath}
\usepackage{ragged2e}
\usepackage{threeparttable} 
\usepackage{url}
\usepackage{xspace}
\usepackage{bbding}

\newcommand{\ie}{\emph{i.e.},\xspace}
\newcommand{\vs}{\emph{v.s.}\xspace}
\newcommand{\eg}{\emph{e.g.},\xspace}

\newcommand\figref[1]{Fig.~\ref{#1}}

\newcommand\tabref[1]{Tab.~\ref{#1}}

\newcommand\secref[1]{Sec.~\ref{#1}}

\newcommand{\fakeparagraph}[1]{\vspace{1mm}\noindent\textbf{#1.}}

\newcommand{\sysname}{{\sf AdaModalX}\xspace}
\newcommand{\sysnameposs}{{\sf AdaModalX's}\xspace}

\newcommand\xxr[1]{\textcolor{black}{#1}}

\begin{document}

\title{\LARGE  AdaModalX: Synchronization-Free MFM Inference for Distributed
Multimodal Mobile Sensing via Progressive Data Orchestration}

\author{Xiangrui Xu, Sicong Liu*, ~\IEEEmembership{Member,~IEEE,} Zhiwen Yu*,~\IEEEmembership{Senior Member,~IEEE,}  \\ Zimu Zhou,~\IEEEmembership{Member,~IEEE,} Bin Guo,~\IEEEmembership{Senior Member,~IEEE} 
\thanks{{*}Corresponding Author: Sicong Liu, Zhiwen Yu (e-mail: scliu@nwpu.edu.cn, zhiwenyu@nwpu.edu.cn)}
\thanks{Xiangrui Xu, Sicong Liu, and Bin Guo with the Northwestern Polytechnical University, Xi'an 710072, China. Zimu Zhou, with the City University of Hong Kong, Hong Kong 999077, China.
Zhiwen Yu with the Northwestern Polytechnical University, Xi'an {\rm 710072} and Harbin Engineering University, Harbin {\rm 150001}, China.}
}

\markboth{Journal of \LaTeX\ Class Files,~Vol.~14, No.~8, August~2021}%
{Shell \MakeLowercase{\textit{et al.}}: A Sample Article Using IEEEtran.cls for IEEE Journals}

\IEEEpubid{0000--0000/00\$00.00~\copyright~2021 IEEE}

\maketitle

\begin{abstract}
\xxr{
Multimodal foundation models (MFMs) support generalizable inference in distributed mobile sensing applications by integrating heterogeneous data streams.
However, existing MFM inference pipelines typically require complete and synchronized multimodal inputs, whereas mobile sensing streams arrive asynchronously due to network variability and heterogeneous data volumes across modalities.
Waiting for all modalities increases end-to-end latency, while inference with incomplete inputs can reduce accuracy.
We argue that multimodal inference should depend on the utility of available information rather than modality completeness.
Based on this principle, we propose \sysname, a progressive multimodal inference framework that decouples inference execution from modality synchronization through runtime utility-aware orchestration.
At each device, importance-aware data uploading uses a lightweight probe distilled from the edge-side MFM to allow reliable samples to exit locally while uncertain samples continue uploading.
At the edge, on-demand MFM activation constructs candidates from fully received samples and, when needed, imputed features derived from partially received delayed inputs.
A lightweight contribution assessor built on frozen MFM encoders estimates candidate utility and activates the MFM once the candidate meets its corresponding utility threshold.
Experiments on four mobile sensing applications show that \sysname improves the accuracy-latency trade-off, achieving up to 26.97\% accuracy improvement and up to 64.75\% end-to-end latency reduction.
}
\end{abstract}


\maketitle

\section{Introduction}
\label{sec:intro}
\begin{figure}[t]
  \centering
  \includegraphics[width=0.50\textwidth]{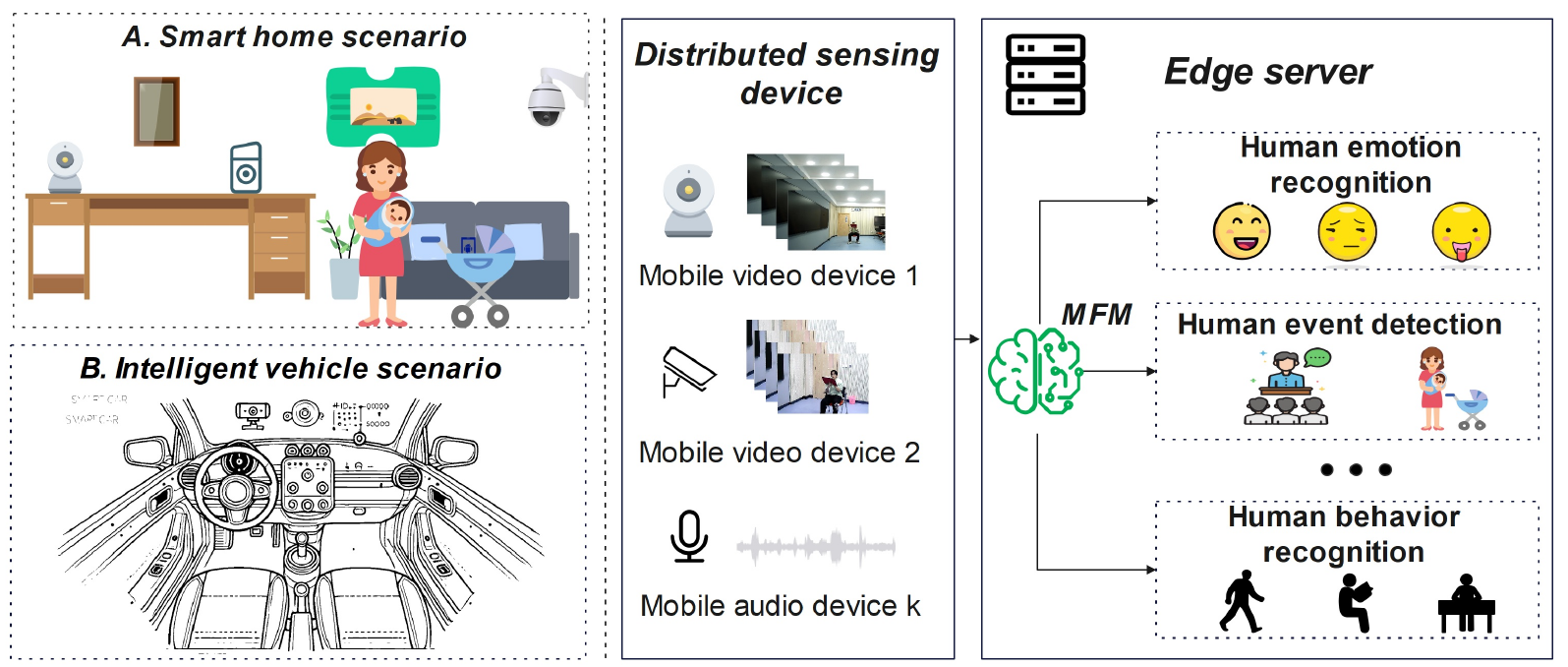}
  \caption{Example distributed mobile sensing applications in smart homes and intelligent vehicles.}
  \vspace{-5pt}
  \label{fig:scenarios}
\end{figure}

Mobile applications increasingly rely on distributed multimodal sensing to support intelligent services and improve user experience~\cite{acmcontext,adaenlight,wang2025towards,cong2026toward}.
These systems employ a network of distributed mobile devices, each equipped with either different or the same sensors, working collaboratively to gather multimodal data streams.
The collected multimodal data are then transmitted to edge or cloud servers, where advanced models perform robust, context-rich inference (see \figref{fig:scenarios}).
For example, in \textit{smart homes}, wearable devices track physiological signals such as heart rate and skin temperature, while ambient sensors monitor environmental factors like humidity and air quality. 
These data streams are processed at the home gateway to detect health anomalies, predict stress levels, or provide personalized wellness insights~\cite{medmllm}. 
Similarly, in \textit{autonomous vehicles}, cameras offer visual scene interpretation, radar identifies objects under varying weather conditions, and LiDAR provides precise spatial mapping. 
These heterogeneous modalities are integrated in \textit{real time} to enable safe navigation and informed decision-making in complex driving environments~\cite{vips}.

These applications are enabled not only by the widespread deployment of mobile and embedded sensors and wireless networks but also by advances in multimodal models for edge-side inference. 
In particular, multimodal deep models~\cite{adaevo} exploit complementary information across heterogeneous sensor streams to support accurate perception and inference.
However, their performance often degrades sharply when deployment involves \textit{zero-shot} or \textit{few-shot} classes, which commonly occurs in mobile environments with evolving users, activities, and contexts.
To address these limitations, multimodal foundation models (MFMs) \cite{mm_llms,jiang2026edge,wei2026survey} have gained prominence for their ability to generalize to  zero- or few-shot classes with little or no fine-tuning \cite{efficient, han2023onellm}. 
This capability supports emerging applications across diverse scenarios, including in-vehicle behavior monitoring (\eg Intel~\cite{mobileye2025}, Tesla~\cite{tesla}), human activity recognition (\eg Microsoft~\cite{babel2025}), and smart-home agents (\eg Amazon~\cite{amazon}, Google Astra~\cite{Google}).

Although MFMs improve the reliability of multimodal inference, their performance is often challenged by \textit{modality asynchrony}, which commonly arises in distributed mobile sensing systems.
In real-world mobile sensing deployments, heterogeneous sensors operate at different sampling rates, generate vastly different data volumes, and experience dynamically varying wireless transmission conditions. 
The resulting modality streams rarely arrive at the edge server simultaneously.
Existing MFMs are largely built upon a \textit{synchronization-dependent execution paradigm}, where inference can proceed only after all modalities become available.
Although recent approaches support flexible modality combinations~\cite{mosel,liu2025deepswarm}, their inference pipelines still rely on \textit{synchronized modality completeness}, making responsiveness fundamentally constrained by the slowest modality stream (see \secref{subsec:asynchrony}).
As a result, waiting for delayed modalities increases inference latency~\cite{MissingDataImputation_li2021low, MissingDataImputationwang2023patch}, while performing inference on incomplete modality states often degrades prediction accuracy due to insufficient cross-modal information~\cite{MissingDataImputationchen2024modality, Data_Augmentation_lee2023multimodal}. 
Existing approaches that reconstruct delayed modalities through raw-data synthesis~\cite{MissingDataImputation_li2021low, MissingDataImputationwang2023patch} further suffer from error propagation across downstream multimodal fusion pipelines.

We argue that multimodal inference should not be treated as a synchronization barrier problem. 
Instead, it should be viewed as a progressive inference process that evolves continuously with modality arrivals.
\IEEEpubidadjcol
Our key insight is that accurate multimodal inference does not always require globally synchronized modality completeness. 
Different modalities contribute unequally to inference outcomes, and delayed
modalities often provide only
marginal incremental utility beyond the information already accumulated 
from currently available streams (see~\secref{subsec:asynchrony}). 
This observation suggests a fundamentally different execution paradigm for distributed multimodal systems, \ie inference should evolve progressively with information accumulation, rather than wait for fully synchronized modality arrivals before execution.

Guided by this principle, we introduce \sysname, a progressive multimodal inference runtime for distributed mobile sensing systems. 
Rather than modifying MFM architectures or retraining large multimodal models into dynamic networks, which demand extensive data and computational resources to preserve their accuracy~\cite{MissingDataImputationlian2023gcnet, JointRepresentation_chen2020hgmf,liu2025crowdhmtware}, \sysname introduces an \textit{orthogonal runtime orchestration layer} that enables existing MFMs~\cite{su2023pandagpt, han2023onellm} to operate efficiently under asynchronous modality arrivals.
\xxr{
Specifically, \sysname first constructs a candidate from all fully received samples and evaluates its utility.
If the candidate does not meet its corresponding utility threshold, it uses the partially received input from the most informative incomplete modality to generate an imputed feature and augment the candidate.
It activates MFM inference once the evaluated candidate meets its corresponding utility threshold, thereby avoiding unnecessary synchronization delays.
}

To realize this capability, we re-purpose frozen MFM encoders with a shared lightweight assessor to learn modality contribution patterns across diverse inputs offline, without altering the original MFM architectures. 
\xxr{
During online execution, \sysname progressively updates the current modality candidate as new modalities arrive and selectively activates MFM inference according to the estimated candidate utility.
}
To further reduce end-to-end latency, \sysname introduces a hierarchical inference architecture spanning mobile devices and the edge server, where lightweight distilled models run locally to process incoming samples and only uncertain samples are offloaded for edge-side MFM inference, thereby significantly reducing wireless transmission overhead and inference latency without sacrificing accuracy.

We implement \sysname and test it across four tasks and four real-world mobile sensing applications.
\xxr{Extensive experiments on four representative scenarios with two MFM inference architectures, four datasets, and four types of mobile devices demonstrate that \sysname consistently outperforms existing synchronous and asynchronous MFM inference methods, achieving superior performance in both accuracy (\eg up to 26.97\% improvement) and responsiveness (\eg up to 64.75\% latency reduction).} 
Notably, these gains are observed under varying modality availability patterns, heterogeneous device conditions, and asynchronous input arrivals, highlighting the robustness and adaptability of our approach. 
By dynamically adapting the inference process to the currently available inputs, \sysname effectively advances the accuracy–latency trade-off frontier in MFM inference.
Our main contributions are summarized as follows.

\begin{itemize}
    \item 
    We identify a fundamental mismatch between synchronization-dependent MFM execution and asynchronous distributed mobile sensing environments, and introduce progressive multimodal inference as a new execution paradigm that evolves with modality arrivals.
    \item 
    \xxr{
    We design \sysname, a progressive multimodal inference runtime that decouples MFM execution from globally synchronized modality completeness.
    It determines whether to exit early based on device-side confidence and, otherwise, when to activate the edge-side MFM according to candidate utility as modality data progressively arrive.
    }
    \item 
    Extensive experiments with four tasks across four devices and dynamic scenarios show that \sysname outperforms existing synchronous/asynchronous methods~\cite{adamml,MissingDataImputation_li2021low,edgefm,shared_specific,liu2025adaflowlite,su2023pandagpt,MissingDataImputationwang2023patch,mosel} in trading off between inference accuracy and low latency across diverse mobile tasks, devices, and scenarios.
\end{itemize}

\section{Motivation}

\label{sec:pre}
\begin{figure}[t]
  \centering
  \subfloat[Task-specific models.]{
    \includegraphics[height=0.165\textwidth]{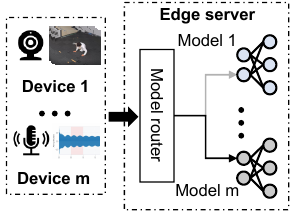}
    \vspace{-5pt}
    \label{fig:taskspecificDNN}
  }
  \subfloat[Task-scalable MFM.]{
    \includegraphics[height=0.165\textwidth]{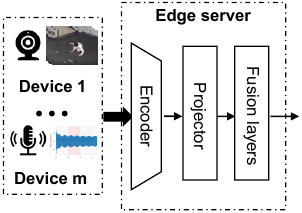}
    \vspace{-5pt}
    \label{fig:scalableMLLM}
  } 
  \caption{Comparison of (a) conventional deep models and (b) task-scalable MFM. The former explicitly selects models based on tasks and input modalities, whereas the latter automatically adapts to tasks and input modalities using the same model.}
  \label{fig:distance_sys}
\end{figure}

\subsection{MFM-based Distributed Mobile Sensing}
Distributed mobile sensing systems have become a core component of mobile computing applications, such as smart homes and autonomous driving systems.
In these systems, distributed devices collaborate to gather multimodal sensory data, which are uploaded to the edge server for accurate and robust inference.
For example,  AliGenie~\cite{AliGenie} collects audio and vision from smartphones, cameras, and other devices,  while edges fuse and analyze these multimodal streams using multimodal deep models (\eg AVEL~\cite{AVE}) to support  mobile applications including emotion recognition~\cite{MOSI}, event detection~\cite{AVE}, and human activity recognition~\cite{driveAct,classter}.

\begin{figure*}[t]
    \centering
    \subfloat[Audio-vision distributed sensing system.]{
      \includegraphics[width=0.42\textwidth]{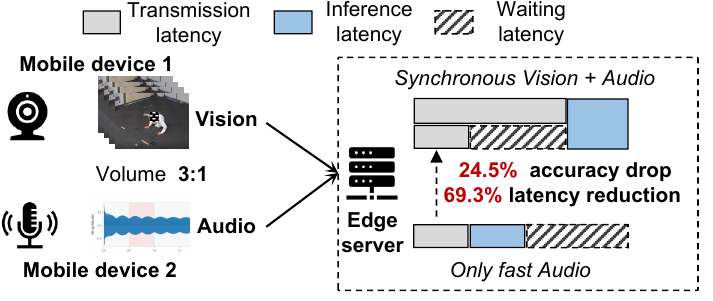}
      \label{fig:asynchronoussystem}
    }
    \subfloat[Few-shot Accuracy.]{
      \includegraphics[width=0.28\textwidth]{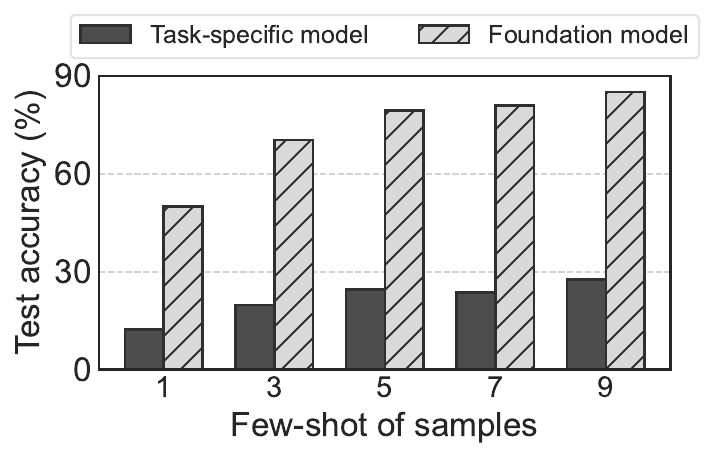}
      \label{fig:ch_1}
    }
    \subfloat[Modality asynchrony.]{
      \includegraphics[width=0.28\textwidth]{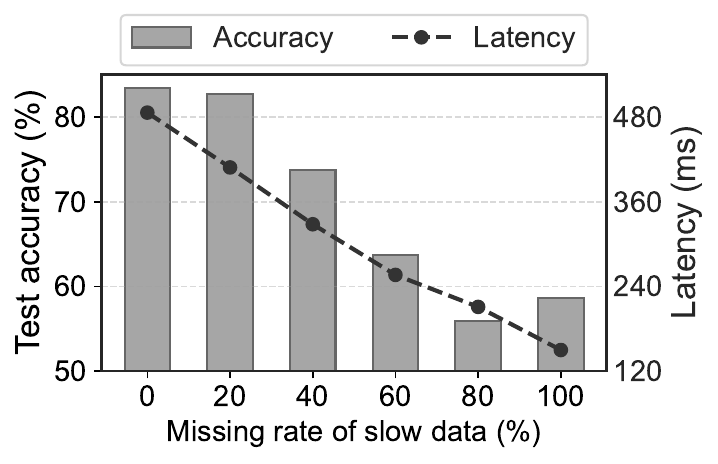}
      \label{fig:ch_2}
    }
    \caption{Impact of modality asynchrony on MFM inference. 
    (a) Illustration of asynchronous data in an audio-vision-based distributed mobile sensing system.
    (b) Inference accuracy with different few-shot settings.
    (c) Inference latency and accuracy with full audio data but partial vision data.}
    \label{fig:ch_all}
\end{figure*}

To scale with the diversity of mobile users and downstream tasks, multimodal foundation models (MFMs) have gained increasing attention for edge-side inference~\cite{efficient}.
Compared to traditional task-specific deep models, MFMs offer the following advantages (see \figref{fig:distance_sys}).
\begin{itemize}
    \item \textbf{Generalization to Diverse Modality Combinations.} 
    MFMs can easily handle diverse input modalities without modifying model architectures~\cite{earthgpt, han2023onellm}.
    For example, in driver behavior recognition, an MFM can integrate RGB, infrared, and depth inputs and adapt to different modality combinations through modality projection~\cite{mm_llms}.
    \item \textbf{Few-shot Generalization to New Environments.} 
    MFMs excel in few-shot and even zero-shot learning~\cite{efficient}, \eg through prompt learning~\cite{few_shot_prompting_1}. 
    For instance, our preliminary experiment on event detection shows  that a pre-trained MFM based on ImageBind~\cite{imagebind} can achieve an accuracy of $70.38\%$, substantially outperforming task-specific models~\cite{adaspring}, which achieve only $19.88\%$ under the 3-shot per-class setting (see \figref{fig:ch_1}).
\end{itemize}

\begin{figure*}[t]
  \centering
  \subfloat[Class-wise performance.]{
    \includegraphics[height=0.165\textwidth]{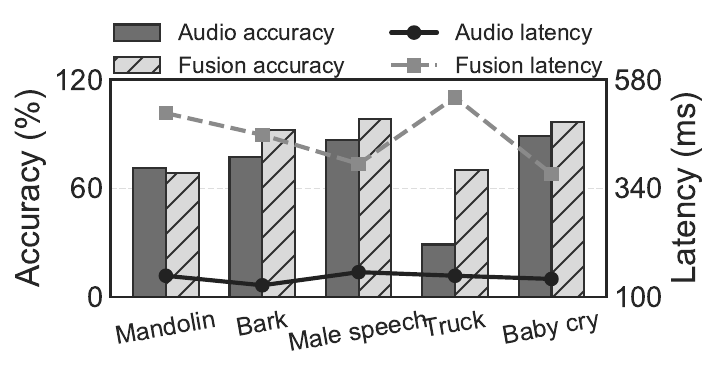}
    \label{fig:Opportunities_class}
  }
  \subfloat[Distribution of exclusive samples.]{
    \includegraphics[height=0.165\textwidth]{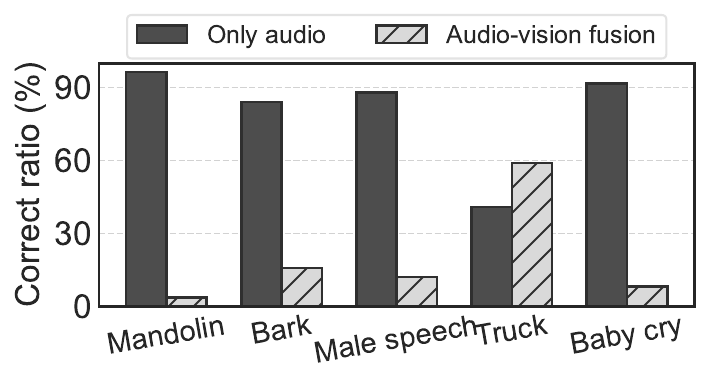}
    \label{fig:Opportunities_sample}
  } 
 \subfloat[Theoretical upper bound.]{
    \includegraphics[height=0.17\textwidth]{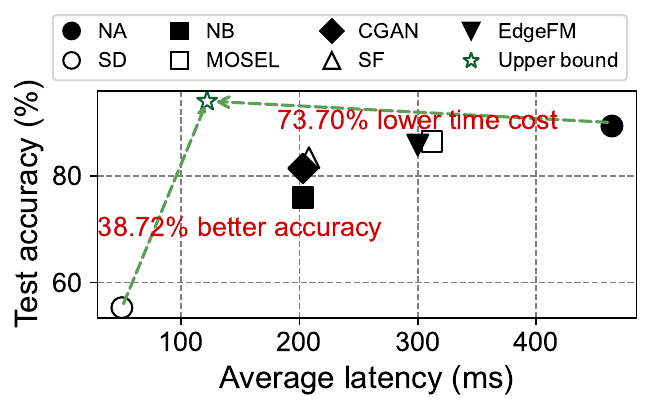}
    \label{fig:Opportunities_upper}
  } 
  \caption{Impact of modalities on different samples. (a) Sample-wise inference accuracy and latency with audio-only \vs both audio and vision inputs. (b) Distribution of mutually exclusive data samples correctly processed by different modalities. (c) Theoretical upper bound of dynamic multimodal inference, illustrating the potential accuracy gain and latency reduction over existing methods.}
  \vspace{-3mm}
  \label{fig:distance_comp}
\end{figure*}

\subsection{MFM Inference with Asynchronous Modalities}
\label{subsec:asynchrony}
\fakeparagraph{Challenges}
Standard multimodal models typically assume that data from all modalities are available for accurate inference \cite{MissingDataImputation_li2021low, MissingDataImputationwang2023patch}.
However, this assumption is often violated in real-world distributed mobile sensing systems.
First, devices often lack synchronization.
Second, transmission delays may vary substantially across modalities because different modalities generate different data volumes. 
For example, under the same bandwidth, vision clips often arrive later than audio because of their larger sizes. 
Consequently, distributed multimodal data streams rarely reach the edge server simultaneously. 
Such modality asynchrony creates a fundamental trade-off between waiting for slower modalities, which increases latency~\cite{MissingDataImputation_li2021low, MissingDataImputationwang2023patch}, and performing early inference with incomplete modalities, which can degrade accuracy~\cite{MissingDataImputationchen2024modality, Data_Augmentation_lee2023multimodal}.

To illustrate this challenge, consider human event detection in a distributed mobile sensing system with audio and vision inputs, where the vision modality arrives later because of its larger size (see \figref{fig:asynchronoussystem}). 
The model uses full audio data and partial vision data for inference.
The missing rate denotes the percentage of vision data not yet arrived when the audio data has been fully received at the edge server. 
If we allow the MFM (ImageBind~\cite{han2023onellm} in our experiment) to perform inference with partial vision data, the inference latency decreases from $0.486s$ to $0.149s$, yet the accuracy also drops drastically from $83.40\%$ to $58.59\%$ on the AVE test dataset \cite{AVE} (see \figref{fig:ch_2}).
The experiment shows that although MFMs can support flexible modality combinations, their accuracy still degrades with (partially) incomplete modalities caused by asynchronously arriving data.

\fakeparagraph{Opportunities}
Inspired by the \textit{conditional computation} principle for efficient inference \cite{han2021dynamic,aiot}, we explore the accuracy-latency trade-off by investigating \textit{whether all modalities are equally important for every sample.}
\figref{fig:Opportunities_class} zooms into the accuracy and latency of five classes when performing MFM inference with either only audio or both audio and vision inputs, using the same model and test dataset as \figref{fig:ch_all}.
We observe that the accuracy of certain classes remains high even when using only audio data (\eg $71.42\%$ with audio only and $68.7\%$ with both audio and vision for the class of Mandolin).
Since inference for these classes does not wait for vision data, its latency is reduced from $0.454s$ to $0.143s$.
\figref{fig:Opportunities_sample} shows similar sample-wise results, where we count the number of samples that can be correctly processed by audio-only or audio-vision fusion.
For example, for the class of Mandolin, $96.33\%$ of samples can be correctly processed with audio only, while merely $3.67\%$ need to be processed with audio-vision fusion.
The above results imply the potential to \textit{dynamically} trigger inference upon receiving a \textit{subset} of \textit{important} multimodal data for \textit{different samples}, rather than \textit{blindly} waiting for \textit{all modalities to arrive}.
Furthermore, prior work~\cite{edgefm} has shown that lightweight on-device models and hierarchical inference can effectively reduce system overhead.

Motivated by this opportunity, we theoretically analyze the optimal execution policy for dynamic MFM inference, as illustrated in \figref{fig:Opportunities_upper}. The analysis shows that compared with existing methods, \textit{progressive multimodal inference} can potentially improve accuracy by $38.72\%$ while reducing latency by $73.70\%$. 
This observation motivates the design of our synchronization-free MFM inference framework via progressive data orchestration for efficient multimodal intelligence.

\section{System Overview}
\label{sec:overview}
This section provides an overview of \sysname, an \textit{orthogonal runtime orchestration layer} for accurate, low-latency MFM inference at the edge, over distributed multimodal inputs arriving asynchronously from mobile devices.
For ease of presentation, we assume that each mobile  device only collects, processes, and sends \textit{single-modality} data to the edge for MFM inference.
Our solution easily applies to mobile devices with multimodal sensors.

\fakeparagraph{Rationales}
As in \secref{subsec:asynchrony},  \sysname improves the accuracy-latency trade-off by \textit{dynamically} triggering MFM inference on a \textit{subset} of \textit{important} modalities rather than waiting for additional delayed modalities.
We realize this idea in distributed mobile sensing systems via the following principles.
\begin{itemize}
    \item \textbf{Progressive Multimodal Inference}.  
    We learn to assess incremental utility for diverse inputs offline by reusing the frozen MFM encoders and training a shared lightweight assessor.
    \xxr{
    During online inference, we trigger MFM inference once the available modalities, including arrived samples and partially received delayed inputs when needed, satisfy semantic sufficiency.
    }
    \item \textbf{Hierarchical Inference}.  
    We utilize mobile devices that can execute lightweight on-device models to perform inference on \textit{local} samples first and only upload \textit{uncertain cases} to the edge for MFM inference.
    This further reduces the overall latency without compromising accuracy. 
\end{itemize}

\begin{figure*}[t]
  \centering
  \includegraphics[width=0.75\textwidth]{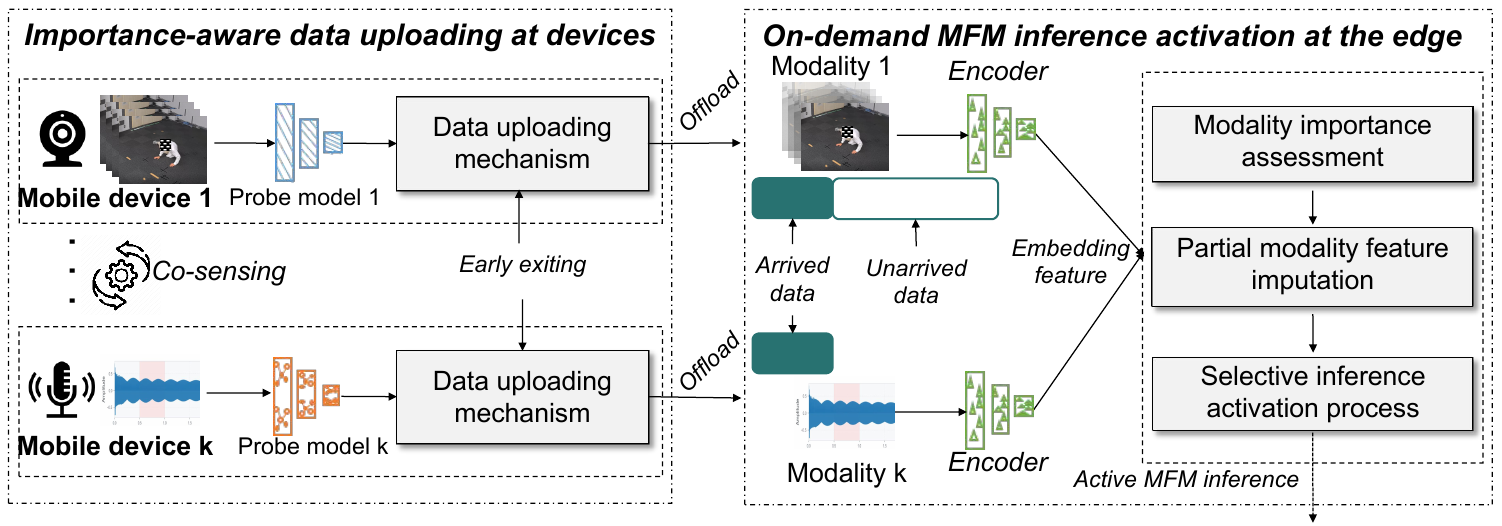}
  \vspace{-5pt}
  \caption{
    Overview of \sysname for MFM-based distributed sensing systems. \xxr{Partial inputs are arrived data available for feature extraction; arrival states indicate per-modality upload progress, not cross-modal data-size correspondence.}
  }
  \vspace{-5pt}
  \label{fig:overview}
\end{figure*}

\fakeparagraph{Workflow}
\sysname consolidates the above rationales into a standard multi-device edge architecture.
\figref{fig:overview} illustrates the workflow of \sysname, which consists of two modules.
\begin{itemize}
    \item \textbf{Importance-aware Data Uploading at Devices} (\secref{sec:AdaFusion}). 
    \xxr{
    Each mobile device runs a lightweight confidence estimator distilled from the edge-side MFM to compute class-wise semantic similarities.
    \sysname derives upload importance from the top-1 similarity and the margin between the top-1 and top-2 similarities.
    Under the joint confidence rule, samples meeting class-wise importance thresholds exit locally; otherwise, uploading continues for edge-side MFM inference.
    }
    \item \textbf{On-demand MFM Inference Activation at Edge} (\secref{sec:non-blocking_inference}). 
    \xxr{
    The edge caches multimodal data asynchronously uploaded from mobile devices and evaluates a candidate formed by fully received modalities using a lightweight contribution assessor.
    If the candidate provides insufficient utility, \sysname imputes a feature from the partially received input of the most informative incomplete modality and re-evaluates the augmented candidate.
    }
    It triggers MFM inference once the candidate satisfies the activation rules.
\end{itemize}
Next, we elaborate on the detailed designs of the two modules.

\section{Importance-aware Data Uploading at Devices}
\label{sec:AdaFusion}

\begin{figure}[t]
  \centering
  \includegraphics[width=0.42\textwidth]{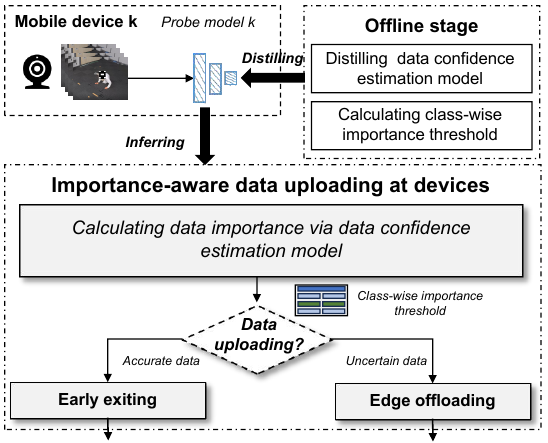}
  \vspace{-5pt}
  \caption{Overview of importance-aware data uploading at devices (\secref{sec:AdaFusion}). 
  }
  \label{fig:Adaptive_data_sampling}
  \vspace{-5pt}
\end{figure}

As shown in \figref{fig:Adaptive_data_sampling}, this module enables each mobile device to decide whether to upload a data sample for edge-side MFM inference using a \textit{confidence-based} importance metric.
It leverages a lightweight \textit{confidence estimation model} (\secref{sec:custom_model}) to obtain sample-wise inference results and confidence locally,  which the \textit{data uploading mechanism} (\secref{sec:data_importance}) then maps to upload importance.
Samples with high confidence exit early from the mobile device, whereas only uncertain samples are uploaded for MFM inference.

\subsection{Data Confidence Estimation}
\label{sec:custom_model}
We develop a lightweight probe model that estimates the inference confidence of each local sample on each device and uses low-confidence evidence as a proxy for upload importance.
To obtain accurate confidence estimates under device resource constraints, we distill the probe from the edge-side MFM.

\fakeparagraph{Model Architecture}
The confidence estimator consists of three modules: a \textit{modality-specific feature extractor}, a \textit{feature projector}, and a \textit{semantic contrastive classifier}.
\begin{itemize}
    \item 
    The feature extractor uses lightweight modality-specific backbones for resource-constrained mobile devices, such as CNN14-based models~\cite{cnn14} and CRNN~\cite{CRNN14} for audio, and ResNet-18~\cite{resnet} and MobileNetV3-Small~\cite{mobilenets,adadeep} for vision.
    \item 
    \xxr{The feature projector aligns embeddings produced by each
modality-specific device-side estimator with the semantic
space of the edge-side MFM,} providing a unified distillation target across heterogeneous devices and modalities.
    \item
    The semantic contrastive classifier computes class-wise semantic similarities and produces both the local prediction and its confidence score.
\end{itemize}

\fakeparagraph{Why Distillation}
We train the confidence estimator by distilling knowledge from the edge-side MFM rather than by training an independent lightweight classifier, for two reasons:
\textit{First}, the estimator is not a standalone classifier; instead, it estimates each sample's \emph{upload importance} for device-side early-exit control.
Distillation from the edge-side MFM provides an \textit{informative decision boundary}, reducing unnecessary uploads while avoiding unreliable early exits.
\textit{Second}, the edge-side MFM provides unified supervision across heterogeneous devices and modalities, whereas each device solves only a simpler local subproblem using its local inputs.
This process transfers shared semantic knowledge to lightweight models, especially when local supervision is scarce.

\fakeparagraph{Confidence-guided Prototype Distillation}
The lightweight estimator jointly produces local predictions and confidence scores to determine whether a sample should exit locally or be uploaded to the edge-side MFM.
Conventional distillation methods~\cite{distilling_1,distilling_2} mainly improve prediction accuracy by transferring logits or feature representations, without explicitly optimizing the confidence margin that governs upload decisions.
This may lead to unnecessary uploads and unreliable early exits.
To address this issue, we introduce a \textit{confidence-guided prototype distillation} loss.

\begin{itemize}
\item
\textbf{Feature-alignment distillation loss.}
Let $z_i$ and $e_i$ denote the device-side and MFM semantic embeddings of sample $x_i$, respectively.
We define the feature-alignment loss as
\begin{equation}
\label{equ:L_FA}
\mathcal{L}_{\text{FA}} = \frac{1}{N} \sum_{i=1}^{N} \|z_i - e_i\|_2^2.
\end{equation}
Minimizing $\mathcal{L}_{\text{FA}}$ aligns device-side embeddings with the MFM semantic space, preserving the discriminative structure of the edge-side MFM.
\item 
\textbf{Pseudo-text contrastive distillation loss.}
\xxr{
Using the MFM embedding, we select the nearest class text prototype as the pseudo target $\tilde{t}_i$ and obtain its confidence $q_i$.
Building on CLIP-style bidirectional contrastive learning~\cite{radford2021learning,imagebind}, we treat $(z_i,\tilde{t}_i)$ as a positive pair and the remaining mini-batch pairs as negatives, and weight the two directional cross-entropy terms by $q_i$, normalized by $\sum_i q_i$, to reduce the effect of uncertain pseudo targets:
}
\begin{equation}
\label{equ:L_PTC}
\begin{aligned}
\mathcal{L}_{\text{PTC}} =
\frac{1}{2} \Bigg(
& \frac{\sum_i q_i \, \ell_{\text{CE}}\!\left(z_i \tilde{T}^{\top}/\tau,\, i\right)}
{\sum_i q_i} \\
& + 
\frac{\sum_i q_i \, \ell_{\text{CE}}\!\left(\tilde{t}_i Z^{\top}/\tau,\, i\right)}
{\sum_i q_i}
\Bigg).
\end{aligned}
\end{equation}
where $z_i$ denotes the device-side embedding, $\tilde{t}_i$ the pseudo-text target, $q_i$ the confidence weight, $Z$ and $\tilde{T}$ the sets of device-side and pseudo-text embeddings, and $\tau$ the temperature.
\item 
\textbf{Class-prototype distillation loss.}
Inspired by cross-modal self-distillation~\cite{Data_Augmentation_li2024unified}, we adopt a cosine-similarity triplet loss to transfer semantic discrimination from text embeddings:
\begin{equation} 
\label{equ:L_CPD}
\begin{aligned}
\mathcal{L}_{\text{CPD}} = \max \big(0, \, 
& \lambda - \text{sim}(z_i, t_i^+) 
+ \text{sim}(z_i, t_i^-) \big),
\end{aligned}
\end{equation}
where $z_i$ denotes the device-side embedding, $t_i^+$ and $t_i^-$ denote the positive and negative class text embeddings, $\text{sim}(\cdot,\cdot)$ denotes cosine similarity, and $\lambda$ is the margin.
\end{itemize}
The overall training objective is
\begin{equation}
\label{equ:L_total}
\mathcal{L}
=
\mathcal{L}_{\text{PTC}}
+
\alpha \mathcal{L}_{\text{FA}}
+
\beta \mathcal{L}_{\text{CPD}},
\end{equation}
where $\alpha$ and $\beta$ are weighting hyperparameters.
We empirically set $\alpha=1$, $\beta=1$, and $\lambda=0.3$ on the AVE dataset~\cite{AVE}.

\fakeparagraph{Confidence-aware Upload Importance Estimation}
\xxr{
Before all modality data arrive at the edge, each device uses its locally captured sample to compute class-wise semantic similarities between the device-side embedding and class prototypes, as defined in Eq. (6).
These similarities provide local confidence estimates for upload decisions, rather than class-wise probabilities over the global multimodal input.
}
Given the confidence estimator, one can quantify inference confidence using an auxiliary model~\cite{BNNConfidence,out_of_dis_detection} or multiple forward/backward passes~\cite{Monte_Carlo_dropout,max_gradient}; however, these methods introduce non-trivial computation and latency overhead.
For rapid confidence estimation on data streams, we follow the softmax-based methods~\cite{appealnet} to quantify inference confidence (and thus data importance). 
 
Specifically, we characterize inference confidence using two complementary metrics: the top-1 similarity score $s_{(1)}(x_i)$, which reflects the \emph{absolute confidence} of the prediction, and the margin score $M_{\text{score}}(x_i)$, which captures \emph{relative class separability}.
The margin score is defined as
\begin{equation}
\label{equ:margin_score_new}
M_{score}(x_i)= s_{(1)}(x_i)-s_{(2)}(x_i),
\end{equation}
where
\begin{equation}
\label{eq6}
s_k(x_i)=\mathrm{sim}(E_d(x_i), c_k), \qquad c_k \in C,
\end{equation}
and $s_{(1)}(x_i)$ and $s_{(2)}(x_i)$ denote the highest and second-highest similarities among all class prototypes.
The margin score captures relative uncertainty, whereas $s_{(1)}(x_i)$ reflects absolute confidence.
Each metric alone is insufficient: the former ignores overall confidence levels, and the latter overlooks inter-class ambiguity.  
Combining them yields a more robust characterization of upload importance.

\subsection{Data Uploading Mechanism}
\label{sec:data_importance}
We therefore design a rule-based decision mechanism that jointly considers absolute confidence and relative class separability and uses \textit{class-wise} importance thresholds to decide whether to upload data or exit early.

\fakeparagraph{Joint confidence rules}
To make upload decisions robust, we jointly evaluate absolute confidence and relative class separability.
Specifically, we define two complementary decision rules:
\begin{itemize}
\item \textbf{High-confidence rule:}
\begin{equation}
\label{high_score}
s_{(1)}(x_i) > \tau^{high}_{top1} \;\;\text{or}\;\; M_{\text{score}}(x_i) > \tau^{high}_{gap},
\end{equation}
\item \textbf{Pairwise-confidence rule:}
\begin{equation}
\label{pair_score}
s_{(1)}(x_i) > \tau^{pair}_{top1} \;\;\text{and}\;\; M_{\text{score}}(x_i) > \tau^{pair}_{gap}.
\end{equation}
\end{itemize}
The device treats sample $x_i$ as \emph{locally reliable} if either rule is satisfied:
\begin{equation}
\text{keep}(x_i) = \mathbb{I}(\text{high-conf} \;\lor\; \text{pair-conf}).
\end{equation}
This hybrid rule improves decision robustness by filtering out low-confidence predictions and boundary-ambiguous samples, thereby reducing unnecessary offloading while avoiding unreliable early exits.

\fakeparagraph{Class-wise Importance Thresholds}
A single global threshold is suboptimal for heterogeneous devices and modalities because their score distributions can differ substantially. 
Recent studies~\cite{edgefm} set device-level thresholds but overlook intra-modality variations across semantic classes.
Using a validation set, we perform an accuracy-constrained search over candidate thresholds derived from sample scores and select the smallest thresholds that satisfy the target accuracy.
We use different targets for the two rules: a higher target for the high-confidence rule to ensure reliability and a lower target for the pairwise-confidence rule to maintain coverage.
To improve robustness, we incorporate two mechanisms:
(i) falling back to global thresholds when class-wise samples are insufficient and (ii) disabling local early exit when no feasible threshold satisfies the target accuracy.

\begin{figure}[t]
  \centering
  \subfloat[Early exit.]{
\includegraphics[height=0.18\textwidth]{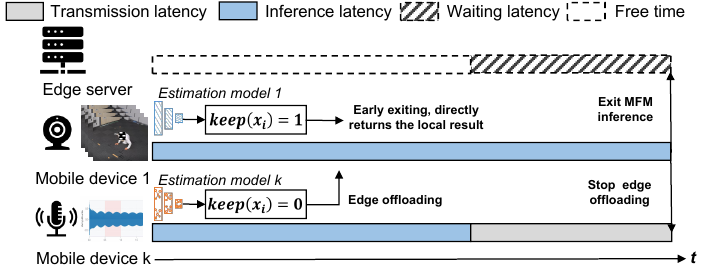}
    \label{fig:exit_data_uploading}
  }
  \\
  \subfloat[Edge offloading.]{
    \includegraphics[height=0.17\textwidth]{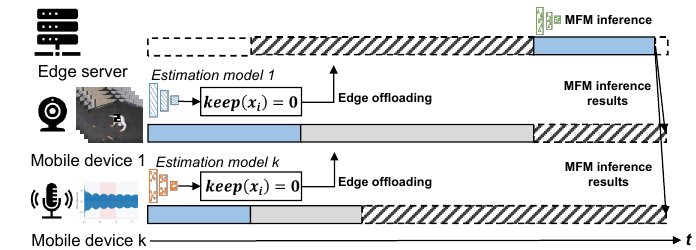}
    \vspace{-5pt}
    \label{fig:continue_data_uploading}
  } 
  \vspace{-5pt}
  \caption{
    Illustration of the confidence-aware uploading mechanism under multi-device coordination.
    Each mobile device performs local probe inference in parallel with data uploading.
    (a) If the estimated confidence is sufficiently high, the device triggers early exit and produces a local prediction.
    (b) If the estimated confidence is insufficient, the device continues uploading, and the edge-side MFM completes the inference.
  }
  \vspace{-5pt}
  \label{fig:decision_data_upload}
\end{figure}
\fakeparagraph{Runtime Data Uploading Decision}
At runtime, each mobile device independently computes $s_{(1)}(x_i)$ and $M_{\text{score}}(x_i)$ from its local input stream. It then applies the calibrated class-wise thresholds to decide whether to exit early or continue uploading, as illustrated in~\figref{fig:decision_data_upload}.
\begin{equation}
\label{equ:decision_new}
\text{decision}_i=
\begin{cases}
\text{Early exiting}, & \text{keep}(x_i)=1,\\
\text{Edge offloading}, & \text{otherwise}.
\end{cases}
\end{equation}
\begin{itemize}
\item \textit{Local early exiting:}
If the confidence rules are satisfied, the device treats its prediction as reliable, stops uploading, and returns the local inference result.
\item \textit{Continued uploading:}
Otherwise, the device continues uploading its data stream for edge-side MFM inference.
\end{itemize}
To minimize latency, we use an opportunistic early-exit strategy:
once any device triggers early exit, the system immediately terminates further uploads and uses that device's prediction as the final result.
If multiple devices satisfy the early-exit condition, the system selects the result from the device that exits first.
This design prioritizes low-latency responses while maintaining reliability through confidence-aware filtering, thereby balancing accuracy and system responsiveness.
\section{On-demand MFM Inference Activation at Edge}
\label{sec:non-blocking_inference}

\begin{figure}[t]
  \centering
  \includegraphics[width=0.50\textwidth]{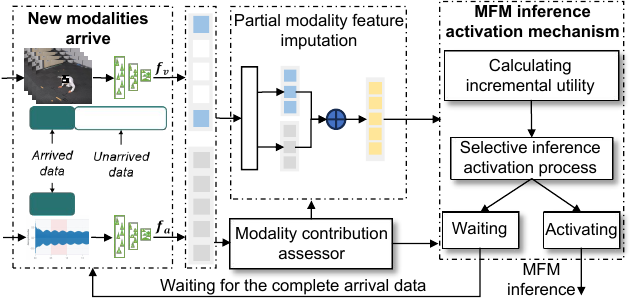}
  \caption{
    Overview of on-demand MFM inference activation at the edge.
  }
  \label{fig:non_blocking_inference}
\end{figure}

This module runs at the edge and determines when to activate MFM inference under asynchronous arrivals of multimodal inputs. 
Let $\{x_1,x_2,\ldots,x_K\}$ denote offloading samples from $K$ modalities.
To avoid blocking inference until all $K$ modalities arrive, \sysname formulates MFM activation as progressive data orchestration.
\xxr{
Upon the complete arrival of a modality sample, the edge evaluates the incremental utility of the available modalities and triggers MFM inference once the utility score of the current candidate meets its candidate-specific activation threshold.
}
This design consists of two modules: a \textit{modality contribution assessor} (\secref{sec:modal_gating}) and an \textit{inference activation mechanism} (\secref{sec:selection}).

\subsection{Modality Contribution Assessor}
\label{sec:modal_gating}
We insert a simple assessor between the encoders and the fusion network and train it to estimate the incremental utility of different modality combinations by learning the discrepancy between a frozen MFM's predictions and the true labels.

\fakeparagraph{Utility Score Definition}
For each modality $k$, the frozen MFM encoder extracts its semantic feature $f_k=E_k(x_k)$.
Given $F_S = \{f_k \mid k \in S\}$, the assessor predicts a utility score $s(F_S) \in [0, 1]$, which estimates whether the candidate feature set contains sufficient information for reliable MFM inference. 
A higher $s(F_S)$ indicates that the candidate set is more informative for the current sample.
\xxr{
Since modality contribution varies across samples and modality combinations, \sysname retains all fully received modality features rather than permanently discarding any modality, while avoiding unnecessary waiting by activating inference once the current candidate meets its utility threshold.
}

\fakeparagraph{Model Architecture}
Since training a separate model for all $2^K-1$ modality subsets is impractical, we introduce a shared lightweight assessment model to estimate the utility of each candidate modality set.
The frozen MFM encoders map heterogeneous modalities into a unified latent space, allowing the same assessment model to evaluate different modality combinations.
We process the candidate feature set $F_S$ with an assessor encoder selected according to the fusion layer of the underlying MFM. 
For LLM-based fusion, we use a lightweight Transformer encoder and take the CLS output as the task-specific semantic representation: $h_S = Trans_\theta(F_S)_{cls}$. 
For MLP-based fusion, we use an MLP encoder: $h_S = MLPEnc_\theta(F_S)$. 
The utility score is then computed as
\begin{equation}
s(F_S) =
\begin{cases}
\sigma (g_\theta(h_S, v_{\mathrm{stat}})), & \text{MLP-based fusion},\\
\sigma (g_\theta(h_S)), & \text{LLM-based fusion}.
\end{cases}
\end{equation}
where $v_{\mathrm{stat}}$ denotes statistical features used only for MLP-based fusion, and $\sigma(\cdot)$ maps the output to $[0,1]$.

\fakeparagraph{Offline Training}
For offline training, we statistically profile the modality arrival order and construct candidate inputs accordingly to train the assessor.
\xxr{
For each candidate feature set $F_S$, we derive the supervisory signal $y_S$ from the predictions of a frozen MFM and the ground-truth labels, and train the contribution assessor by minimizing
$
L_{\mathrm{imp}}=\sum_S\bigl(s(F_S)-y_S\bigr)^2.
$
We formulate the selection of candidate-specific activation thresholds as an accuracy-constrained latency minimization problem.
Let $\mathrm{Acc}_{F_S}(V)$ and $\mathrm{Lat}_{F_S}(V)$ denote the validation accuracy and average latency under the activation rule $s(F_S)\geq V$, respectively.
The threshold for $F_S$ is selected by solving
\[
V_{F_S}
=
\arg\min_{V\in\mathcal{V}_{F_S}}
\mathrm{Lat}_{F_S}(V),
\quad
\mathrm{s.t.}\quad
\mathrm{Acc}_{F_S}(V)\geq A_{\mathrm{tar}},
\]
where $\mathcal{V}_{F_S}$ contains the candidate thresholds derived from validation-set utility scores, and $A_{\mathrm{tar}}$ is the target accuracy.
We solve this problem by grid search and select the feasible threshold with the minimum average latency.
During inference, \sysname triggers MFM inference once the current candidate satisfies $s(F_S)\geq V_{F_S}$.
}

\xxr{
For dynamic arrival orders and modality subsets whose candidate feature sets $F_S$ are covered during offline training, \sysname reuses the shared contribution assessor and corresponding $V_{F_S}$ without online retraining, reselecting only the thresholds when deployment conditions change.
For an unseen combination of MFM-supported modalities, it keeps the MFM frozen, updates only the assessor, and selects the corresponding $V_{F_S}$ on a validation set.
An MFM-unsupported modality additionally requires an encoder or projector to map it into the MFM semantic space before the assessor is updated and the corresponding thresholds are selected offline.
}

\begin{figure*}[t]
  \centering
  \includegraphics[width=0.80\textwidth]{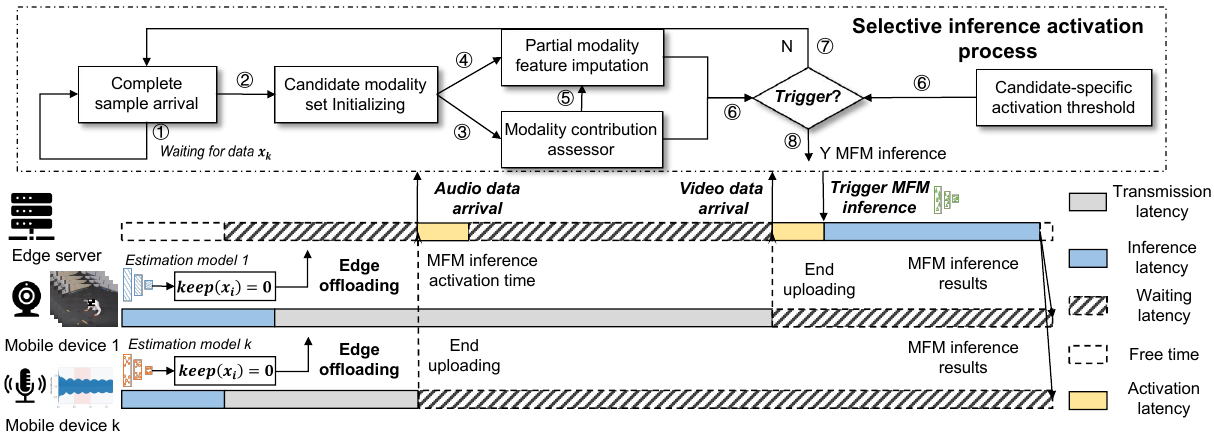}
  \vspace{-5pt}
  \caption{
  Overview of the MFM inference activation mechanism. Mobile devices $1$ to $k$ use a confidence-aware uploading mechanism to continuously upload multimodal data.
  Once the audio sample arrives completely (\textcircled{1}), it triggers MFM inference activation (\textcircled{2}), calculates the modality utility via assessor (\textcircled{3}), uses partial-modality feature imputation to recover complementary multimodal information (\textcircled{4}), predicts the imputed feature (\textcircled{5}), and combines them into a candidate set to check if the utility meets its \xxr{candidate-specific activation threshold} (\textcircled{6}). 
  If the candidate set does not meet the threshold (\textcircled{7}), \sysname waits for the next data arrival (\textcircled{1}); otherwise, it triggers MFM inference (\textcircled{8}).
  }
  \label{fig:inference_activation}
  \vspace{-5pt}
\end{figure*}

\subsection{MFM Inference Activation Mechanism}
\label{sec:selection}
Instead of relying on complete modality arrival synchronization~\cite{adamml} or strict maximum latency deadlines~\cite{mosel}, we progressively evaluate whether to execute MFM inference or await further inputs whenever a new modality is fully received.

For ease of presentation, we assume that the samples ${x_k}$ arrive at the edge in ascending order; \ie the complete sample $x_k$ (\eg an audio frame) is received earlier than $x_{k+1}$ (\eg a video clip).
\xxr{
Once $x_k$ fully arrives, \sysname updates the current candidate $\mathcal{F}_k$ with the features of $\{x_1,\ldots,x_k\}$.
If $k=K$, \sysname directly runs MFM inference.
Otherwise, it evaluates $\mathcal{F}_k$ and triggers inference if it meets its activation rule.
If $\mathcal{F}_k$ fails its threshold, \sysname combines the partial feature $f_p^{part}$ of the most informative incomplete modality with $\mathcal{F}_k$ to impute $\hat{f}_p$ and evaluates the augmented candidate $\widetilde{\mathcal{F}}_k$.
\sysname waits for the next modality only if the augmented candidate also fails its threshold.
}

\sysname implements the above decision process with two optimizations: \textit{(i)} partial modality feature imputation and \textit{(ii)} a selective inference activation process, as explained next.

\fakeparagraph{Partial Modality Feature Imputation}
Rather than relying only on fully received modality inputs or waiting for the remaining modality samples to arrive completely, we exploit partially received modalities via \textit{feature-level imputation}.
For the most informative incomplete modality $x_p \in {x_{k+1},\ldots,x_K}$, we encode its partially received input into a feature $f_p^{\mathrm{part}}$ and impute the missing task-relevant semantics conditioned on the available feature set $\mathcal{F}_k$.
The resulting augmented candidate $\tilde{\mathcal{F}}_k=\mathcal{F}_k\cup{\hat{f}_p}$ enables earlier MFM activation while reducing accuracy degradation due to incomplete modality arrivals.

Specifically, \sysname performs feature imputation in the shared semantic space of the MFM, rather than reconstructing the raw missing data~\cite{adaflow}.
Given the available feature set $\mathcal{F}_k={f_1,\ldots,f_k}$ and the partial feature $f_p^{\mathrm{part}}=E_p(x_p^{\mathrm{part}})$ produced by the modality encoder, we extract compact cross-modal context tokens as
\begin{equation}
    C_k = \mathrm{Ctx}_{\theta}(\mathcal{F}_k \cup \{f_p^{\mathrm{part}}\}),
\end{equation}
where $\mathrm{Ctx}_{\theta}(\cdot)$ is a modality-aware token compressor that projects heterogeneous modality features into a unified hidden space and summarizes them through lightweight token mixing and latent-query cross-attention.
Conditioned on the partial feature $f_p^{\mathrm{part}}$ and the compact context $C_k$, the model generates a task-relevant semantic residual:
\begin{equation}
    \Delta f_p = G_{\theta}(f_p^{\mathrm{part}}, C_k), \quad
    \hat{f}_p = f_p^{\mathrm{part}} + \Delta f_p .
\end{equation}
Here, $G_{\theta}(\cdot)$ is implemented as a lightweight cross-attention imputer followed by an MLP residual head.
By adding only the predicted residual, the imputation preserves the semantic evidence already contained in the partial feature while recovering complementary information for early MFM activation.
In the offline stage, we freeze the MFM and train the imputation module using the following objective:
$
\mathcal{L}=\mathcal{L}_{\mathrm{task}}+\lambda \mathcal{L}_{\mathrm{reg}},
$
where $\lambda$ is set to $0.2$, and $\mathcal{L}_{\mathrm{reg}}$ denotes the regularization loss.
Here, $\mathcal{L}_{\mathrm{task}}$ is the task-specific loss (\eg an MLLM uses the causal language modeling loss).

\textbf{Selective Inference Activation Process.}
When $x_k$ is fully received, the modality samples $\{x_1,\ldots,x_k\}$ are complete, whereas $\{x_{k+1},\ldots,x_K\}$ remains incomplete.

\begin{itemize}
\item AdaModalX incrementally encodes the newly arrived sample $x_k$ into $f_k$ and updates the cached feature set as $\mathcal{F}_k=\mathcal{F}_{k-1}\cup\{f_k\}$.
\xxr{
If $k=K$, all modality samples have arrived, and AdaModalX directly performs MFM inference with $\mathcal{F}_K$.
Otherwise, it first evaluates the candidate formed by fully received modalities.
If this candidate satisfies the activation rule, i.e., $s(\mathcal{F}_k)\ge V_{\mathcal{F}_k}$, AdaModalX triggers MFM inference with $\mathcal{F}_k$.
}
\item If the complete candidate set is insufficient, AdaModalX selects the most informative incomplete modality $x_p$ from $\{x_{k+1},\ldots,x_K\}$ for imputation rather than simply waiting for the next modality sample to become complete.
AdaModalX encodes the partial modality sample $x_p^{part}$ into a partial feature $f_p^{part}$.
The imputation module then combines $f_p^{part}$ with $\mathcal{F}_k$ to generate an augmented feature $\hat{f}_p$.
The assessor scores the augmented candidate $\widetilde{\mathcal{F}}_k=\mathcal{F}_k\cup\{\hat{f}_p\}$ to make the next activation decision.
If this candidate satisfies the activation rule, i.e., $s(\widetilde{\mathcal{F}}_k)\ge V_{\widetilde{\mathcal{F}}_k}$, AdaModalX triggers MFM inference with $\widetilde{\mathcal{F}}_k$; otherwise, it waits until the next modality sample from the mobile client is fully received at the edge.
\end{itemize}

\section{Experiment}
\subsection{Experiment Setup}
\fakeparagraph{Implementation}
We implement \sysname in Python 3.12.3 and PyTorch 2.5.1 for the server and mobile devices.
The edge server is equipped with an RTX 3090 GPU, Intel(R) Xeon(R) Gold 6133 CPU @ 2.50GHz, and 256GB of RAM.
\xxr{
We evaluate AdaModalX on four mobile devices: Jetson Orin NX ($D_1$), Jetson Orin Nano ($D_2$), Xiaomi 13 Ultra ($D_3$), and Vivo X300 Pro ($D_4$).
}
To simulate the distributed multimodal asynchronous arrival problem, we also evaluate the system's performance across varying network bandwidths.
For importance-aware data uploading (\secref{sec:AdaFusion}), we use a shallow MLP for text and lightweight CNNs for audio and vision as data-confidence estimators. 
\xxr{
We repeat each experiment five times with different random seeds and report the mean $\pm$ standard deviation. For key comparisons with small performance differences, we additionally conduct statistical significance tests, with the results indicated in the corresponding figures.
}

\fakeparagraph{Task, Datasets, and Model}
We implement two representative MFM inference architectures to evaluate system performance:
\begin{itemize}
\item \textbf{MFM-based Inference} ($M_1$): 
This architecture consists of a multimodal encoder, a projector, a fusion module, and a contrastive classifier. 
It maps inputs into a unified space, fuses them, and matches the fused representation against class text embeddings for classification. 
We implement it using ImageBind~\cite{imagebind}. 
For example, in event detection tasks, the prompt template is: ``Event is $CLS$.''
\item \textbf{MLLM Inference} ($M_2$): 
This architecture uses the complete MLLM to directly output the final class label texts from multimodal inputs and task prompts. 
We implement it using Qwen3.5-2B~\cite{qwen35blog}. For example, in multimodal sentiment analysis~\cite{MOSI}, the prompt is: ``You are a multimodal sentiment classification expert. Based on the input, classify the sentiment as positive, neutral, or negative. The sentiment is ().''
\end{itemize}

We further evaluate \sysname on four representative real-world distributed multimodal applications:
\begin{itemize}
\item \textbf{Event Detection} ($T_1$):
\xxr{Audio-visual event recognition on the AVE dataset~\cite{AVE}, which contains 4,143 videos across 28 event categories; we divide each video into 1-second clips for recognition.
}
\item \textbf{Driver Activity Recognition} ($T_2$):
Multimodal driver behavior understanding on drive\&Act~\cite{driveAct}, a multi-view, multi-modal driver activity dataset.
\item \textbf{Human Activity Recognition} ($T_3$): Human action recognition on the Kinetics benchmark~\cite{kay2017kinetics}, a large-scale video-audio dataset for action classification.
\item \textbf{Human Sentiment Analysis} ($T_4$):
\xxr{
Sentiment classification on CMU-MOSI~\cite{MOSI}, which contains aligned text, audio, and visual modalities.
}
\end{itemize}

\fakeparagraph{Baseline}
We compare \sysname with \textit{nine} representative baselines that span both synchronous and asynchronous multimodal inference. 
\xxr{
For fair comparison, we reproduce all baseline methods in the same evaluation pipeline and evaluate them under the same settings. 
The baselines include:
}
\begin{itemize}
\item \textbf{Synchronous MFM inference}:
    The inference pipeline waits for all input data to arrive before the server performs MFM inference on the complete modalities.
    \begin{itemize}
    \item \textbf{No-adaptation (NA)}~\cite{imagebind,qwen35blog}:
    The edge server waits for all modalities to arrive and then runs the MFM inference using all modalities.
    \item \textbf{Modality Gating (MG)}~\cite{adamml,dynamic_multimodal_fusion, smg}:
    It uses a lightweight policy network to dynamically select input modalities for MFM inference, thereby reducing latency and computational overhead.
    \item \textbf{Minimal Modality Inference (Min)}~\cite{architecting}:
    Triggers inference immediately with the fastest-arriving modality (\eg audio), prioritizing early inference but possibly sacrificing accuracy.
    \end{itemize}
\item \textbf{Asynchronous MFM inference}~\cite{MissingDataImputationwang2023patch}:
   The MFM inference pipeline can perform inference when partial multimodal data arrive. 
   \begin{itemize}
   \item\textbf{Original Non-Blocking (NB)}~\cite{MissingDataImputationwang2023patch}:
   It directly runs MFM inference on the currently available partial multimodal input once one modality fully arrives.
    \item \textbf{Standalone (SD)}~\cite{resnet}: This runs a lightweight single-modal model, distilled from the MFM, exclusively on mobile devices for local inference.
    \item \textbf{MOSEL}~\cite{mosel}: It dynamically selects a subset of available modalities for MFM inference to meet latency and accuracy requirements.
    \item \textbf{CGAN-based Modality Imputation (CGAN)}~\cite{MissingDataImputation_li2021low,liu2025adaflowlite, MissingDataImputationwang2023patch,adaflow}: It uses a CGAN to impute slow modality features, thereby improving inference accuracy.
    \item \textbf{Shared Feature (SF)}~\cite{shared_specific}: It uses a specific- and shared-feature network to reconstruct missing modality features for asynchronous MFM inference.
    \item \textbf{EdgeFM}~\cite{edgefm}: It selectively uploads uncertain samples to edge-side FM and performs model switching according to data uncertainty and network conditions.
    \end{itemize}
\end{itemize}

\subsection{Performance Comparison}

\begin{figure*}[t]
 \raggedright
  \centering
  \subfloat[$T_{1}$, $M_{1}$, 10Mbps.]{
    \includegraphics[height=0.18\textwidth]{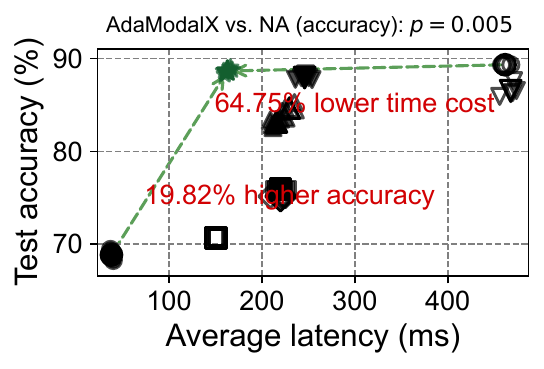}
    \vspace{-5pt}
    \label{fig:ave_m1_all}
  }
  \subfloat[$T_{2}$, $M_{1}$, 50Mbps.]{
    \includegraphics[height=0.18\textwidth]{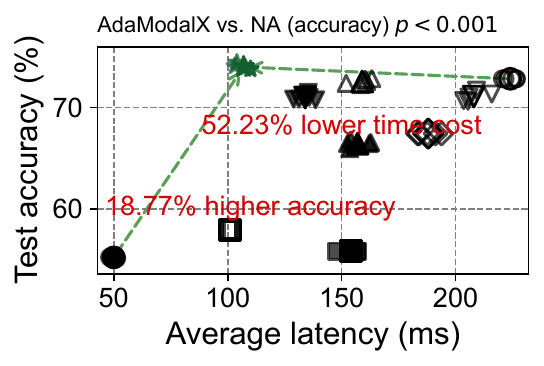}
    \vspace{-5pt}
    \label{fig:drive_m1_all}
  }
  \subfloat[$T_{3}$, $M_{1}$, 2Mbps.]{
    \includegraphics[height=0.18\textwidth]{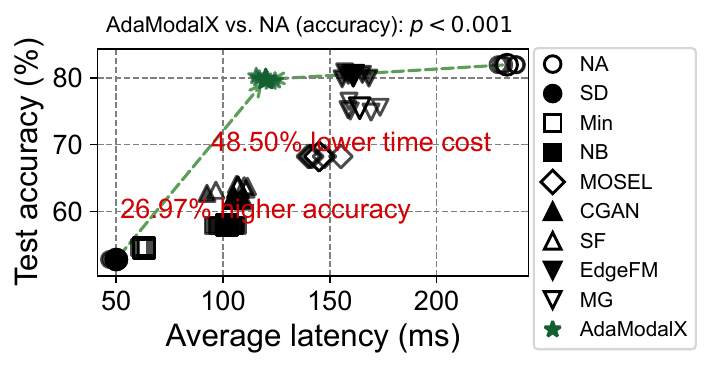}
    \vspace{-5pt}
    \label{fig:kinetics_m1_all}
  }
  \
  \subfloat[$T_{4}$, $M_{1}$, 15Mbps.]{
    \includegraphics[height=0.18\textwidth]{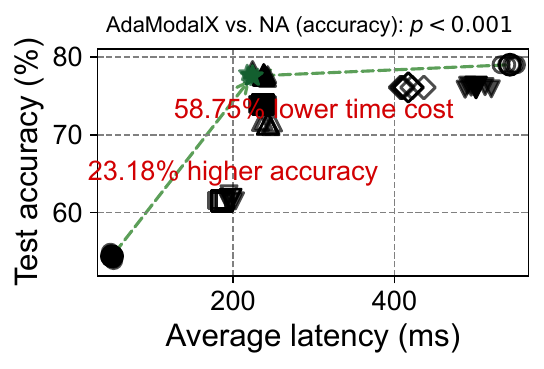}
    \vspace{-5pt}
    \label{fig:mosi_m1_all}
  }
  \subfloat[$T_{1}$, $M_{2}$, 10Mbps.]{
    \includegraphics[height=0.18\textwidth]{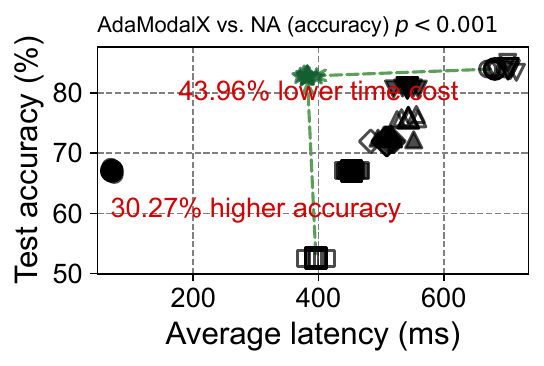}
    \vspace{-5pt}
    \label{fig:M2_AVE_overall}
  }
  \subfloat[$T_{3}$, $M_{2}$, 2Mbps.]{
    \includegraphics[height=0.18\textwidth]{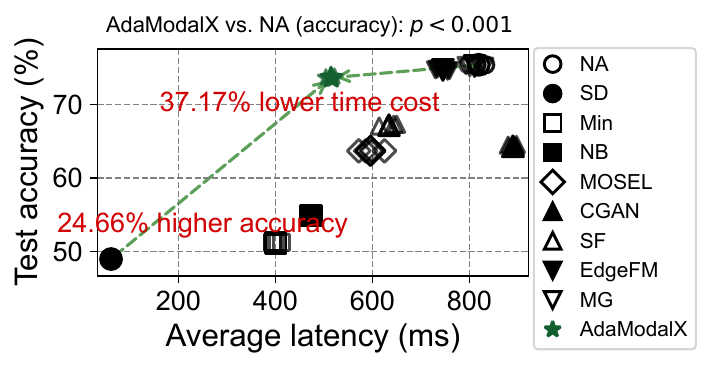}
    \vspace{-5pt}
    \label{fig:M2_kinetics_overall}
  }
  \vspace{-5pt}
  \caption{\xxr{Comparison of accuracy \vs latency between AdaModalX and other baselines on diverse tasks. Translucent markers show the five individual runs, while opaque markers show their means. The annotations report paired-test $p$-values for the accuracy comparison between AdaModalX and NA.}}
  \vspace{-5pt}
  \label{fig:acc_let_total}
\end{figure*}

\begin{figure*}[t]
  \centering
  \subfloat[Event detection ($T_1$), $M_{1}$.]{
    \includegraphics[height=0.15\textwidth]{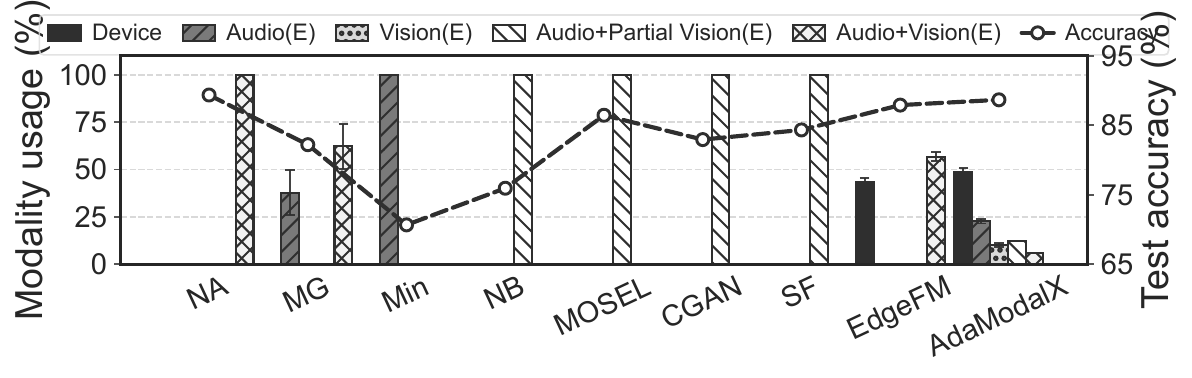}
    \label{fig:ave_modality_use}
  }
  \subfloat[Driver activity recognition ($T_2$), $M_{1}$.]{
    \includegraphics[height=0.15\textwidth]{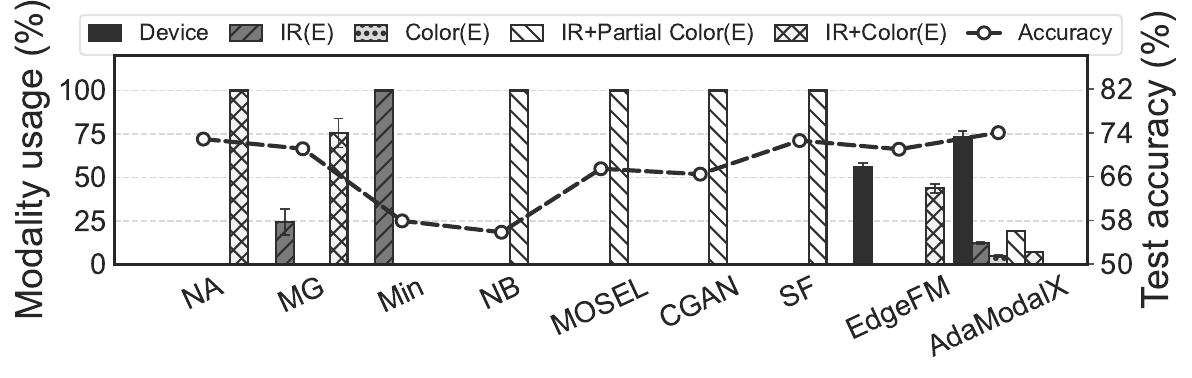}
    \label{fig:drive_modality_use}
  }
  \vspace{-5pt}
  \caption{ Comparison of modality usage rate(\%) \vs accuracy(\%) between \sysname and other approaches on diverse tasks.}
  \vspace{-8pt}
  \label{fig:modality_use}
\end{figure*}

\begin{figure}[t]
  \centering
  \subfloat[$T_1$, $M_{1}$, 10Mbps.]{
    \includegraphics[height=0.14\textwidth]{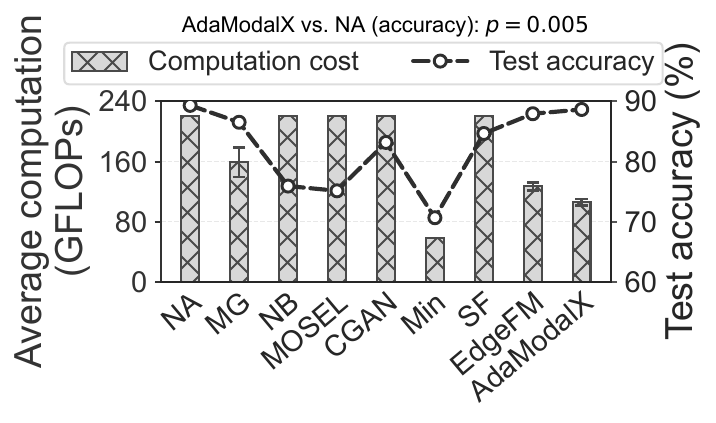}
    \label{fig:ave_com_cost}
  }
  \subfloat[$T_2$, $M_{1}$, 20Mbps.]{
    \includegraphics[height=0.14\textwidth]{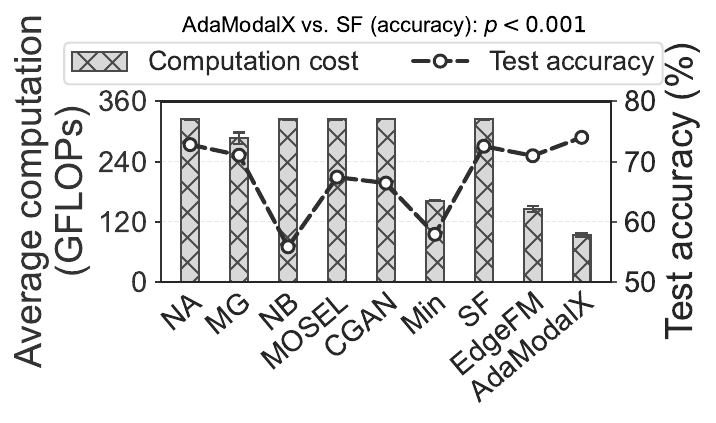}
    \label{fig:drive_com_cost}
  }
  \caption{\xxr{Comparison of computational cost between AdaModalX and other approaches on diverse tasks. The annotations report paired-test $p$-values for AdaModalX versus NA in (a) and AdaModalX versus SF in (b).}}
  \vspace{-5pt}
  \label{fig:com_cost}
\end{figure}

\subsubsection{System Latency \vs Accuracy}
We evaluate the inference accuracy and system latency of \sysname under different network conditions across four tasks ($T_1$--$T_4$) and two multimodal foundation models ($M_1$~\cite{imagebind} and $M_2$~\cite{qwen35blog}), comparing it with nine baselines (NA~\cite{imagebind}, MG~\cite{adamml}, Min~\cite{architecting}, NB~\cite{MissingDataImputationwang2023patch}, SD~\cite{resnet}, MOSEL~\cite{mosel}, CGAN~\cite{adaflow}, SF~\cite{shared_specific}, and EdgeFM~\cite{edgefm}).
\textit{First}, as in \figref{fig:acc_let_total}, \sysname achieves the best accuracy-latency trade-off across tasks, models, and network conditions. 
Compared with NA~\cite{imagebind}, it reduces the average latency by 50.89\% over the four tasks and two models, with less than 2\% accuracy degradation. 
Compared with Min~\cite{architecting}, it improves the average accuracy by 21.52\%.
\textit{Second}, it achieves the lowest system latency while maintaining competitive accuracy. 
For example, as shown in \figref{fig:drive_m1_all}, on the $T_2$ task under 50Mbps network bandwidth, it reduces latency by 52.23\%, 48.56\%, 30.52\%, 43.09\%, 31.85\%, 32.70\%, and 20.15\% compared with NA~\cite{imagebind}, MG~\cite{adamml}, NB~\cite{MissingDataImputationwang2023patch}, MOSEL~\cite{mosel}, CGAN~\cite{adaflow}, SF~\cite{shared_specific}, and EdgeFM~\cite{edgefm}, respectively. 
It also improves accuracy by 16.07\% and 18.77\% compared with Min~\cite{architecting} and SD~\cite{resnet}, respectively.

\textbf{Summary.} 
\sysname achieves the best accuracy-latency trade-off in  distributed asynchronous multimodal sensing via importance-aware uploading and on-demand activation.

\subsubsection{Modality Usage}
We evaluate the modality usage of \sysname under different network bandwidths for the event detection and driver activity recognition tasks, comparing it with baselines that require data uploading for edge-side inference, including NA~\cite{imagebind}, MG~\cite{adamml}, Min~\cite{architecting}, NB~\cite{MissingDataImputationwang2023patch}, MOSEL~\cite{mosel}, CGAN~\cite{adaflow}, SF~\cite{shared_specific}, and EdgeFM~\cite{edgefm}.
\textit{First}, \sysname achieves the lowest edge-side modality usage among all methods.
As shown in \figref{fig:drive_modality_use}, 73.31\% of samples are inferred directly on mobile devices, improving the early-exit ratio by 17.10\% compared with EdgeFM~\cite{edgefm} by avoiding unnecessary data uploading.
\textit{Second}, \sysname significantly reduces the usage of slow modalities. Compared with MG~\cite{adamml} and NA~\cite{imagebind} in \figref{fig:modality_use}, it reduces the usage of the slow modality by 53.76\% and 71.25\%, respectively.
\textit{Third}, it also reduces the usage of the IR-Color modality combination by 93.11\% compared with NA~\cite{imagebind} and reduces the usage of the IR-Partial Color modality combination by 80.83\% compared with NB~\cite{MissingDataImputationwang2023patch}, MOSEL~\cite{mosel}, CGAN~\cite{adaflow}, and SF~\cite{shared_specific}.

\textbf{Summary.}
\sysname reduces modality usage and latency through early exiting at devices and on-demand activation at the edge while preserving competitive accuracy.

\subsubsection{Computational Cost}
We further evaluate the computational cost and inference accuracy of \sysname using $M_1$ on the event detection ($T_1$) and driver activity recognition ($T_2$) tasks under 10Mbps and 20Mbps, comparing it with eight baselines: NA~\cite{imagebind}, MG~\cite{adamml}, Min~\cite{architecting}, NB~\cite{MissingDataImputationwang2023patch}, MOSEL~\cite{mosel}, CGAN~\cite{adaflow}, SF~\cite{shared_specific}, and EdgeFM~\cite{edgefm}.
\textit{First}, \sysname achieves the lowest computational cost while maintaining competitive accuracy across different tasks, as shown in \figref{fig:com_cost}.
For example, in \figref{fig:drive_com_cost}, \sysname reduces the computational cost by 71.31\%, 67.52\%, 71.31\%, 71.31\%, 71.31\%, 42.61\%, 71.31\% and 36.11\% compared with NA~\cite{imagebind}, MG~\cite{adamml}, NB~\cite{MissingDataImputationwang2023patch}, MOSEL~\cite{mosel}, CGAN~\cite{adaflow}, Min~\cite{architecting}, SF~\cite{shared_specific}, and EdgeFM~\cite{edgefm}, respectively.
It improves accuracy by 2.97\%, 16.11\%, 18.18\%, 6.61\%, 7.60\%, and 1.48\% compared with MG~\cite{adamml}, Min~\cite{architecting}, NB~\cite{MissingDataImputationwang2023patch}, MOSEL~\cite{mosel}, CGAN~\cite{adaflow}, and SF~\cite{shared_specific}, respectively.

\textbf{Summary.} 
\sysname effectively reduces computational costs while maintaining competitive accuracy via importance-aware uploading and on-demand inference activation.

\begin{figure*}[t]
  \centering
  \subfloat[$T_2$, $M_{1}$, Accuracy.]{
    \includegraphics[height=0.15\textwidth]{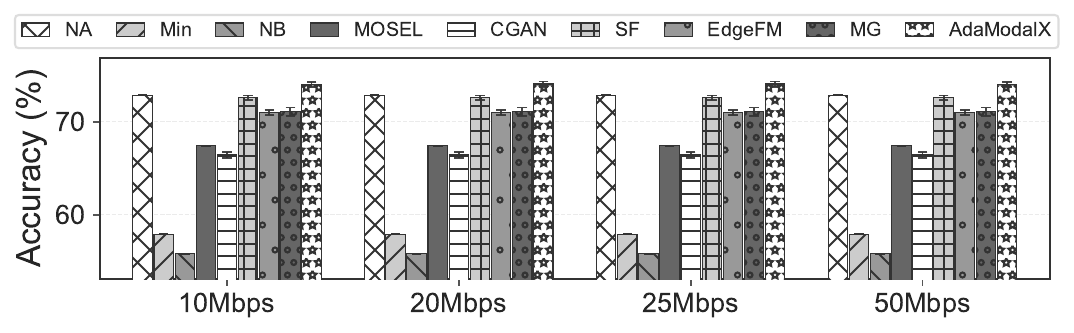}
    \label{fig:ave_band_cost}
  }
  \subfloat[$T_2$, $M_{1}$, Latency.]{
    \includegraphics[height=0.15\textwidth]{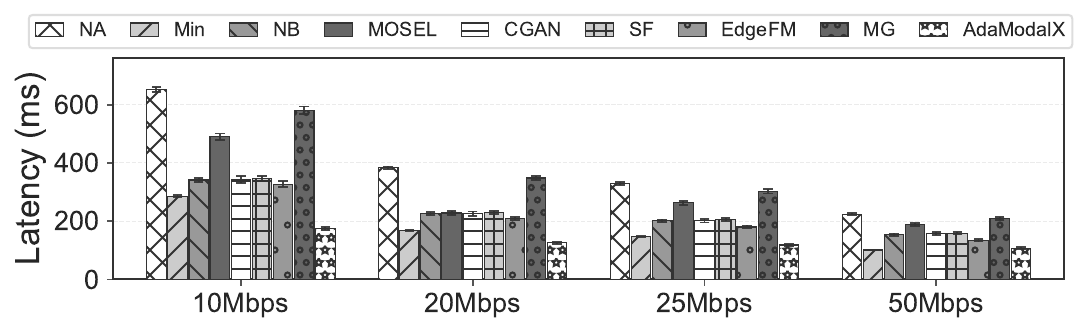}
    \label{fig:drive_band_cost}
  }
  \vspace{-5pt}
  \caption{\xxr{Bandwidth sensitivity of \sysname on $T_2$ with $M_1$ under 10Mbps, 20Mbps, 25Mbps, and 50Mbps.}}
  \vspace{-5pt}
  \label{fig:band_cost}
\end{figure*}

\begin{table}[t]
\tiny
\centering
\caption{Ablation study on importance-aware data uploading at devices (\#D1) and on-demand inference at edge (\#D2) on $T_1$ and $T_2$ tasks under different bandwidths.}
\label{tab:ablation_design}
\begin{tabular}{|c|c|c|c|c|}
\hline
\multirow{2}{*}{\textbf{Method}} 
& \multicolumn{2}{c|}{\textbf{Event detection ($T_1$), 10 Mbps}} 
& \multicolumn{2}{c|}{\textbf{Driver activity recognition ($T_2$), 20 Mbps}} \\
\cline{2-5}
& \textbf{Accuracy (\%)} 
& \textbf{Latency (ms)}
& \textbf{Accuracy (\%)} 
& \textbf{Latency (ms)} \\
\hline

\textbf{w/o \#D1, w/o \#D2} 
& 89.35 $\pm$ 0.00 
& 462 $\pm$ 4.05 
& 72.82 $\pm$ 0.00
& 383 $\pm$ 3.72 \\
\hline

\textbf{w/ \#D1, w/o \#D2} 
& 87.93 $\pm$ 0.25
& 258 $\pm$ 5.94
& 73.44 $\pm$ 0.32
& 179 $\pm$ 4.18 \\
\hline

\textbf{w/o \#D1, w/ \#D2} 
& \textbf{88.73 $\pm$ 0.63}
& 234 $\pm$ 2.86
& \textbf{75.25 $\pm$ 0.41}
& 247 $\pm$ 3.67 \\
\hline

\textbf{w/ \#D1, w/ \#D2} 
& 88.66 $\pm$ 0.28
& \textbf{163 $\pm$ 3.93}
& 73.99 $\pm$ 0.27
& \textbf{107 $\pm$ 3.26} \\
\hline

\end{tabular}
\end{table}

\subsubsection{\xxr{Bandwidth Sensitivity}}
We evaluate system latency and accuracy on the driver activity recognition task ($T_2$) under different network bandwidths (\ie 10Mbps, 20Mbps, 25Mbps, and 50Mbps), as shown in \figref{fig:band_cost}.
\xxr{First, \sysname maintains competitive accuracy while consistently reducing latency across different network bandwidths.}
Compared with NA~\cite{imagebind}, \sysname reduces latency by 73.16\%, 67.10\%, 64.13\%, and 52.23\% under 10Mbps, 20Mbps, 25Mbps, and 50Mbps, respectively.
\textit{Second}, \sysname shows greater latency advantages as the network bandwidth decreases.
For example, when the bandwidth decreases from 50Mbps to 10Mbps, the latency reduction of \sysname over NA~\cite{imagebind}, Min~\cite{architecting}, NB~\cite{MissingDataImputationwang2023patch}, MOSEL~\cite{mosel}, CGAN~\cite{adaflow}, SF~\cite{shared_specific}, MG~\cite{adamml}, and EdgeFM~\cite{edgefm} further increases by 20.93\%, 44.75\%, 18.16\%, 21.13\%, 17.13\%, 16.72\%, 21.27\%, and 26.33\%, respectively.

\textbf{Summary.} 
\xxr{\sysname maintains competitive accuracy across bandwidth settings, with larger latency advantages under bandwidth-constrained networks.}

\subsection{Understanding \sysnameposs Improvement}
\label{sec:data_resampling}

\subsubsection{Impact of Design 1 (\secref{sec:AdaFusion}) and Design 2 (\secref{sec:non-blocking_inference})}
In \tabref{tab:ablation_design}, we evaluate the impact of importance-aware data uploading at devices (\#D1) and on-demand MFM inference activation at the edge (\#D2) on the $T_1$ and $T_2$ tasks using $M_1$.
\textit{First}, enabling \#D1 alone reduces latency by 48.71\% on average while maintaining competitive accuracy, indicating that on-device early exit effectively avoids unnecessary data uploads for both $T_1$ and $T_2$.
\xxr{
\textit{Second}, enabling \#D2 alone reduces latency by 42.43\% on average, suggesting that on-demand MFM inference activation mitigates waiting for slow modalities.
\textit{Third}, jointly enabling \#D1 and \#D2 reduces latency by 68.39\% on average while maintaining competitive accuracy, demonstrating the complementarity of the two designs for distributed multimodal mobile sensing.
}

\begin{figure*}[t]
  \centering
  \subfloat[Event detection ($T_1$), $M_{1}$.]{
    \includegraphics[height=0.17\textwidth]{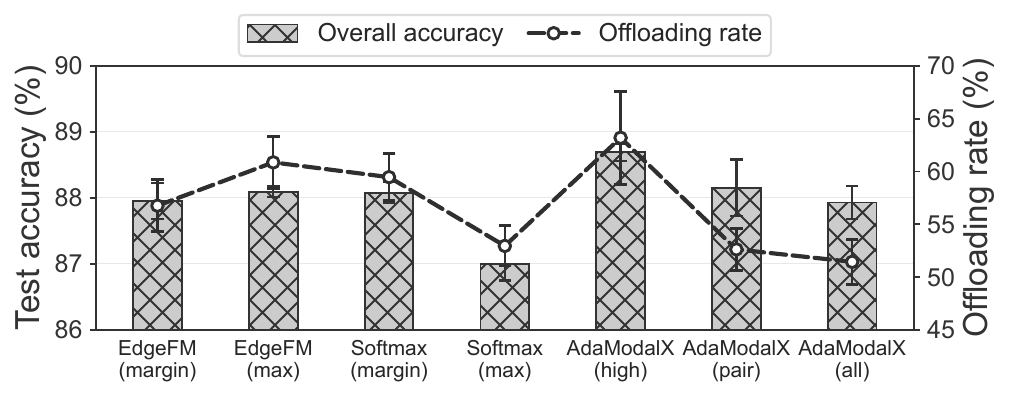}
    \label{fig:d1_ave}
  }
  \subfloat[Driver activity recognition ($T_2$), $M_{1}$.]{
    \includegraphics[height=0.17\textwidth]{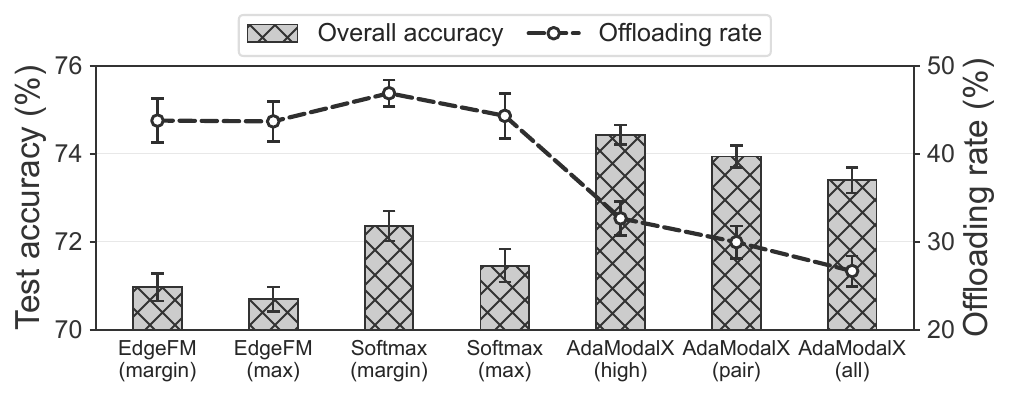}
    \label{fig:d1_drive}
  }
  \caption{Comparison of accuracy \vs offloading rate(\%) between importance-aware data uploading at devices (\#D1) and six baselines on $T_1$ and $T_2$ tasks.}
  \vspace{-5pt}
  \label{fig:adamodalx_1}
\end{figure*}

\begin{figure*}[t]
 \raggedright
  \subfloat[$T_{1}$, $M_{1}$, 10Mbps.]{
    \includegraphics[height=0.18\textwidth]{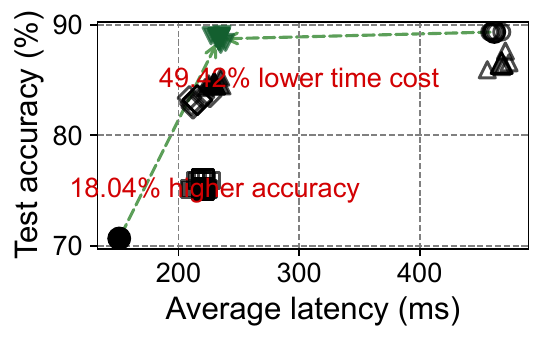}
    \vspace{-5pt}
    \label{fig:ave_m1_d2}
  }
  \subfloat[$T_{2}$, $M_{1}$, 50Mbps.]{
    \includegraphics[height=0.18\textwidth]{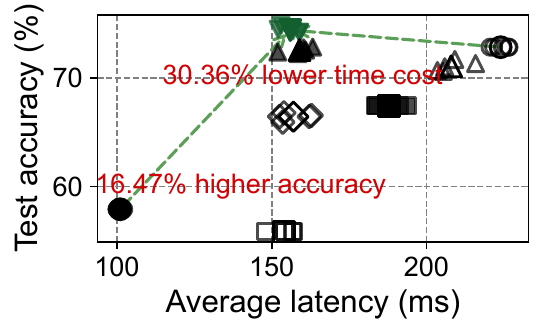}
    \vspace{-5pt}
    \label{fig:drive_m1_d2}
  }
  \subfloat[$T_{3}$, $M_{1}$, 2Mbps.]{
    \includegraphics[height=0.18\textwidth]{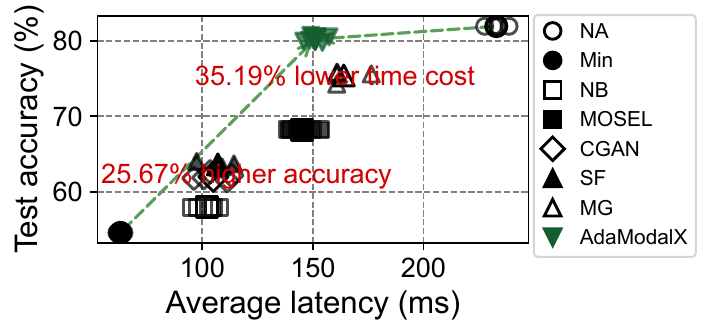}
    \vspace{-5pt}
    \label{fig:kinetics_m1_d2}
  }
  \caption{\xxr{Effectiveness of on-demand MFM inference activation at edge (\#D2) with the importance-aware data uploading at devices (\#D1) disabled.}}
  \vspace{-5pt}
  \label{fig:acc_let_d2}
\end{figure*}
\subsubsection{Performance of Importance-aware Data Uploading at Devices}
We test overall accuracy and offloading rate of the device-side importance-aware data uploading module (\#D1) on the $T_1$ and  $T_2$ tasks using $M_1$ and compare it with EdgeFM(margin)~\cite{edgefm}, EdgeFM(max)~\cite{score_1}, Softmax(margin)~\cite{edgefm}, Softmax(max)~\cite{score_1}, \sysname(high), and \sysname(pair). 
Here, margin uses the gap between the top-1 and top-2 scores as the confidence metric, while max directly uses the top-1 score.
\sysname(high) and \sysname(pair) use only the high-confidence rule or the pairwise-confidence rule.
As shown in \figref{fig:adamodalx_1}, \#D1 achieves the lowest offloading rates of 51.44\% and 26.68\% on $T_1$ and $T_2$, respectively, while maintaining competitive overall accuracy.
Specifically, compared with EdgeFM\textit{(margin)}~\cite{edgefm}, EdgeFM\textit{(max)}~\cite{score_1}, Softmax\textit{(margin)}~\cite{edgefm}, Softmax\textit{(max)}~\cite{score_1}, \sysname\textit{(high)}, and \sysname\textit{(pair)}, \#D1 reduces the offloading rate by 17.10\%, 17.01\%, 20.21\%, 17.64\%, 5.99\%, and 3.29\% on $T_2$, respectively.

\begin{figure*}[t]
  \centering
  \subfloat[Event detection ($T_1$), $M_{1}$, arriving rate=25\%.]{
    \includegraphics[height=0.17\textwidth]{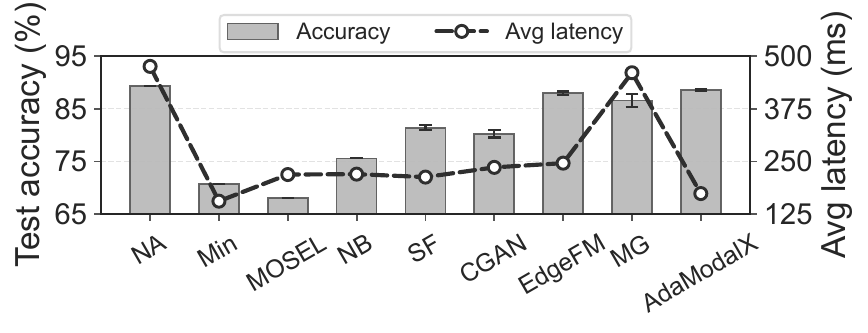}
    \label{fig:75_missing}
  }
  \subfloat[Event detection ($T_1$), $M_{1}$, arriving rate=50\%.]{
    \includegraphics[height=0.17\textwidth]{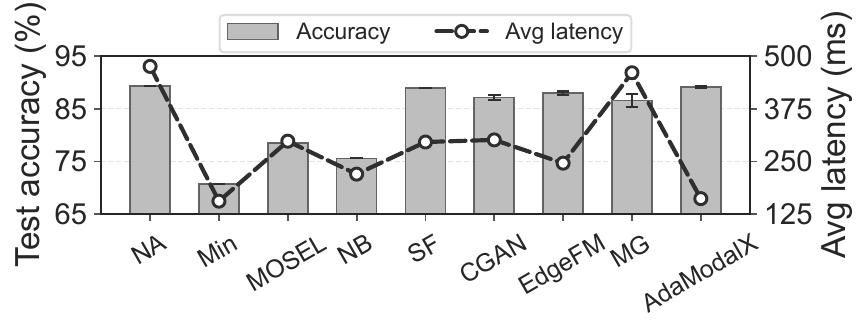}
    \label{fig:50_missing}
  }
  \caption{\xxr{Sensitivity to slow-modality arrival rates on $T_1$ with $M_1$.}}
  \vspace{-5pt}
  \label{fig:missing_rate}
\end{figure*}
\subsubsection{Performance of On-demand MFM Inference Activation at Edge}
We evaluate the accuracy and latency of the edge-side on-demand MFM inference activation module (\#D2) on  $T_1$, $T_2$, and $T_3$ tasks under diverse bandwidths, compared with seven baselines (NA~\cite{imagebind}, Min~\cite{architecting}, NB~\cite{MissingDataImputationwang2023patch}, MOSEL~\cite{mosel}, CGAN~\cite{adaflow}, SF~\cite{shared_specific}, and MG~\cite{adamml}).
\xxr{As shown in \figref{fig:acc_let_d2}, \#D2 independently reduces waiting latency while maintaining competitive accuracy across tasks and bandwidth settings.}
Compared with NA~\cite{imagebind}, it reduces the average latency by 37.76\% over the three tasks, with competitive accuracy, and it improves the average accuracy by 20.06\% compared with Min~\cite{architecting}.
\xxr{
For example, in \figref{fig:drive_m1_d2}  for $T_2$ under 50Mbps, \#D2 reduces latency by 30.36\%, 17.02\%, and 25.0\% compared with NA~\cite{imagebind}, MOSEL~\cite{mosel}, and MG~\cite{adamml}, respectively.
}

\subsubsection{\xxr{Sensitivity to Slow-modality Arrival Rates}}
In \figref{fig:missing_rate}, we evaluate the performance of \sysname under different asynchronous settings and compare it with eight baselines (NA~\cite{imagebind}, MG~\cite{adamml}, Min~\cite{architecting}, NB~\cite{MissingDataImputationwang2023patch}, MOSEL~\cite{mosel}, CGAN~\cite{adaflow}, SF~\cite{shared_specific}, and EdgeFM~\cite{edgefm}) on the $T_1$ task using $M_1$.
We vary the arrival rate of the slow modality (vision), defined as the proportion of vision data that has arrived when the audio fully arrives.
\xxr{
At the 50\% arrival rate, it reduces latency by 65.9\% over
NA~\cite{imagebind} while maintaining competitive accuracy,
and reduces latency by 45.3\% over the strongest non-NA
baseline SF~\cite{shared_specific}.
Second, \sysname remains effective under a lower arrival rate.
At the 25\% arrival rate, it reduces latency by 63.4\% over
NA~\cite{imagebind} while maintaining competitive accuracy.
These results demonstrate the robustness of \sysname to
different degrees of modality asynchrony.
}

\subsection{Micro-benchmarks and Ablation Studies}

\begin{table}[t]
\scriptsize
\centering
\caption{Ablation study on different distillation loss components in the IR modality.}
\label{tab:loss_ablation_ir}
\resizebox{\columnwidth}{!}{%
\begin{tabular}{|c|c|c|c|}
\hline
\textbf{Distillation loss}
& \textbf{Model acc. (\%)}
& \textbf{Overall acc. (\%)}
& \textbf{Offloading rate (\%)} \\
\hline
\begin{tabular}[c]{@{}c@{}}
w/ $\mathcal{L}_{PTC}$, w/o $\mathcal{L}_{FA}$,
w/o $\mathcal{L}_{CPD}$
\end{tabular}
& 63.55 $\pm$ 0.56
& 74.03 $\pm$ 0.42
& 46.67 $\pm$ 2.14 \\
\hline
\begin{tabular}[c]{@{}c@{}}
w/ $\mathcal{L}_{PTC}$, w/ $\mathcal{L}_{FA}$,
w/o $\mathcal{L}_{CPD}$
\end{tabular}
& 63.77 $\pm$ 0.44
& 74.39 $\pm$ 0.37
& 44.87 $\pm$ 1.86 \\
\hline
\begin{tabular}[c]{@{}c@{}}
w/ $\mathcal{L}_{PTC}$, w/o $\mathcal{L}_{FA}$,
w/ $\mathcal{L}_{CPD}$
\end{tabular}
& 63.23 $\pm$ 0.52
& 74.39 $\pm$ 0.41
& 44.51 $\pm$ 2.48 \\
\hline
\begin{tabular}[c]{@{}c@{}}
w/ $\mathcal{L}_{PTC}$, w/ $\mathcal{L}_{FA}$,
w/ $\mathcal{L}_{CPD}$
\end{tabular}
& 67.69 $\pm$ 0.36
& 75.70 $\pm$ 0.54
& 38.61 $\pm$ 1.63 \\
\hline
\end{tabular}%
}
\end{table}

\subsubsection{Impact of Different Distillation Loss}
We evaluate the impact of different distillation loss components on the $T_2$ task by mobile model accuracy, overall accuracy, and offloading rate.
W/ $\mathcal{L}_{PTC}$, w/ $\mathcal{L}_{FA}$, and w/ $\mathcal{L}_{CPD}$ achieves the lowest offloading rate while maintaining competitive accuracy in \tabref{tab:loss_ablation_ir}.
Compared with the configurations w/ $\mathcal{L}_{PTC}$, w/o $\mathcal{L}_{FA}$, w/o $\mathcal{L}_{CPD}$; w/ $\mathcal{L}_{PTC}$, w/ $\mathcal{L}_{FA}$, w/o $\mathcal{L}_{CPD}$; and w/ $\mathcal{L}_{PTC}$, w/o $\mathcal{L}_{FA}$, w/ $\mathcal{L}_{CPD}$, the full configuration (w/ $\mathcal{L}_{PTC}$, w/ $\mathcal{L}_{FA}$, and w/ $\mathcal{L}_{CPD}$) reduces the offloading rate by 8.06\%, 6.26\%, and 5.9\%, respectively.


\begin{figure*}[t]
  \centering
  \subfloat[Threshold $\tau^{high}_{top1}$.]{
    \includegraphics[height=0.16\textwidth]{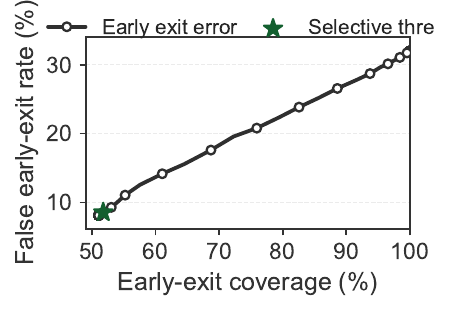}
    \label{fig:high_top1}
  }
  \subfloat[Threshold $\tau^{high}_{gap}$.]{
    \includegraphics[height=0.16\textwidth]{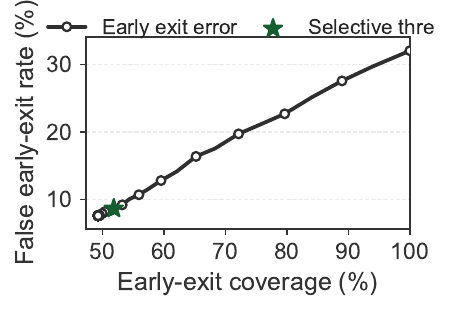}
    \label{fig:high_gap}
  }
  \subfloat[Threshold $\tau^{pair}_{top1}$.]{
    \includegraphics[height=0.16\textwidth]{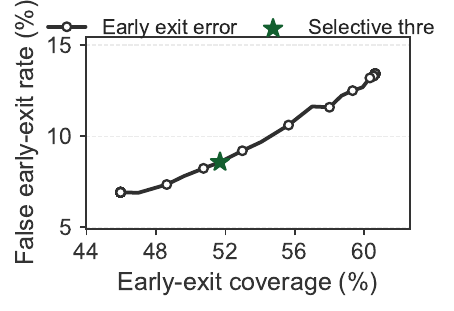}
    \label{fig:pair_top1}
  }
  \subfloat[Threshold $\tau^{pair}_{gap}$.]{
    \includegraphics[height=0.16\textwidth]{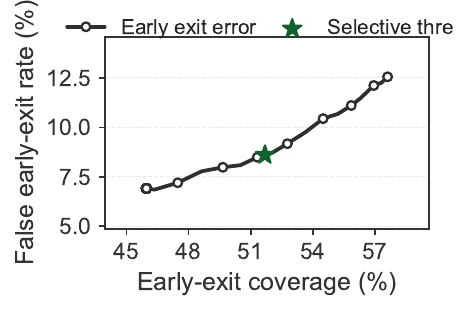}
    \label{fig:pair_gap}
  }
  \vspace{-5pt}
  \caption{\xxr{Reliability of false early-exit rate (FEER) and early-exit coverage under four device-side confidence thresholds on $T_1$ with $M_1$.}}
  \vspace{-5pt}
  \label{fig:thre}
\end{figure*}

\begin{figure}[t]
  \centering
  \subfloat[Similarity margin.]{
    \includegraphics[height=0.17\textwidth]{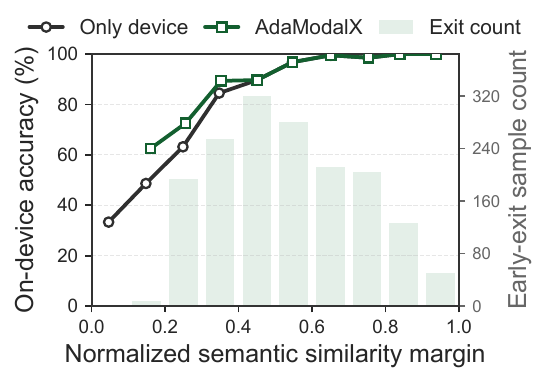}
    \label{fig:top1_top2_accuracy_curve}
  }
  \subfloat[Maximum  similarity.]{
    \includegraphics[height=0.17\textwidth]{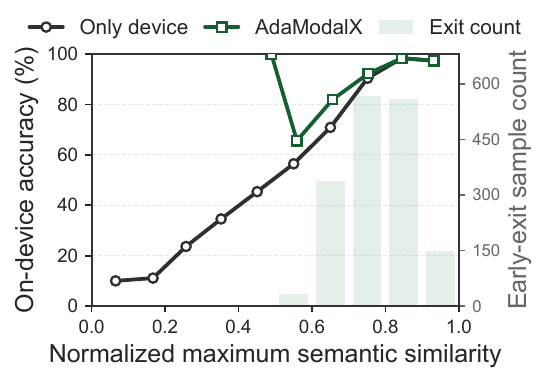}
    \label{fig:top1_accuracy_curve}
  }
  \caption{\xxr{Reliability of device-side early exit under two confidence scores.}}
  \vspace{-5pt}
  \label{fig:thre_1}
\end{figure}

\begin{table}[t]
\centering
\caption{\xxr{Overall device-side early-exit outcomes of \sysname on $T_1$ with $M_1$.}}
\label{tab:early_exit_reliability}
\scriptsize
\setlength{\tabcolsep}{4pt}
\renewcommand{\arraystretch}{1.15}
\begin{tabular}{|c|c|c|c|}
\hline
\textbf{Method} &
\shortstack{\textbf{Overall early-exit}\\\textbf{rate (\%)}} &
\shortstack{\textbf{Overall false early-exit}\\\textbf{rate (\%)}} &
\shortstack{\textbf{Overall edge rescue}\\\textbf{rate (\%)}} \\
\hline
\sysname & 52.62 & 4.24 & 2.69 \\
\hline
\end{tabular}
\vspace{-2mm}
\end{table}

\begin{table}[t]
\centering
\scriptsize
\caption{Ablation study on the effects of partial multimodal feature imputation (I) and selective inference activation (S) on the accuracy and latency of event detection.}
\label{tab:ablation}
\begin{tabular}{|c|c|c|}
\hline
\multirow{2}{*}{\textbf{Method}}
& \multicolumn{2}{|c|}{\textbf{Event detection}} \\
\cline{2-3}
& \textbf{Average accuracy (\%)} & \textbf{Average latency (ms)} \\
\hline
w/o I, w/o S & 75.95 $\pm$ 0.00 & 220 $\pm$ 5.20 \\
\hline
w/ I, w/o S  & 83.53 $\pm$ 0.60 & 223 $\pm$ 0.41 \\
\hline
w/o I, w/ S  & 86.25 $\pm$ 0.51 & 254 $\pm$ 6.70 \\
\hline
w/ I, w/ S   & 88.73 $\pm$ 0.63 & 244 $\pm$ 2.86 \\
\hline
\end{tabular}
\end{table}

\subsubsection{Benefits of on-demand MFM inference activation at edge}
We evaluate the benefits of on-demand MFM inference activation at the edge by ablating partial multimodal feature imputation (I) and selective inference activation (S) on $T_1$ tasks. 
As in~\tabref{tab:ablation}, compared with w/o I and w/o S, w/ I, w/o S improves accuracy by 7.58\%.
Enabling inference activation yields a larger accuracy gain of 10.30\%.
The w/ I and w/ S setting achieves the best accuracy, improving accuracy by 12.78\%.


\subsubsection{\xxr{Reliability of Device-Side Early Exit}}
\xxr{
We evaluate the reliability of the device-side confidence-aware early-exit rule on $T_1$ with $M_1$.
\textit{First}, as shown in \figref{fig:thre}, we evaluate the trade-off between the false early-exit rate (FEER) and early-exit coverage, i.e., early-exit rate, by varying each of the four confidence thresholds in the hybrid rule while fixing the other three. Across all four threshold sweeps, FEER generally increases with early-exit coverage, while the selected thresholds achieve a balanced FEER--coverage trade-off.
\textit{Second}, as shown in \figref{fig:thre_1}, we group test samples into score intervals to assess the empirical reliability of the two semantic confidence scores.
The early-exited samples achieve comparable or higher on-device accuracy in most populated score intervals, indicating that the hybrid rule retains a more reliable subset of device-side predictions.
\textit{Third}, \tabref{tab:early_exit_reliability} quantifies the false early-exit outcomes under the selected threshold setting. \sysname early-exits 52.62\% of test samples, while only 4.24\% of all samples are false early exits; among all samples, 2.69\% correspond to false early exits that would have been corrected by the edge model if uploaded. This shows that \sysname reduces edge inference while limiting unrecovered local errors.
}

\begin{figure}[t]
  \centering
  \includegraphics[width=0.45\textwidth]{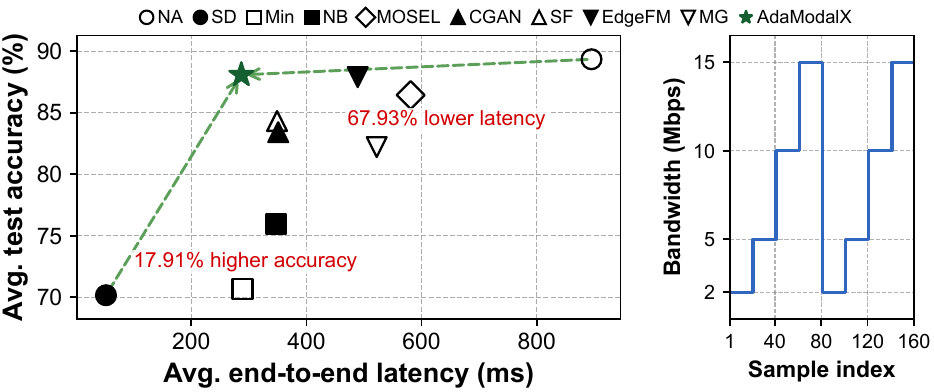}
  \vspace{-3mm}
  \caption{\xxr{Comparison of accuracy \vs latency under network fluctuation on  $T_1$ with $M_1$.}}
  \label{fig:network_fluctuation}
  \vspace{-3mm}
\end{figure}

\subsubsection{\xxr{Performance under bandwidth fluctuations}}
\xxr{
We evaluate \sysname on $T_1$ with $M_1$ using a sample-sequence bandwidth trace that cycles through $\{2,5,10,15\}$\,Mbps every 20 samples.
As shown in~\figref{fig:network_fluctuation}, \sysname achieves the best accuracy-latency trade-off.
Compared with NA~\cite{imagebind}, it reduces the average latency by 67.93\% while maintaining competitive accuracy.
At comparable latency to Min~\cite{architecting}, it improves accuracy by 17.91\%.
These results demonstrate the robustness of \sysname to bandwidth fluctuations in modality transmission delays.
}

\begin{table*}[t]
\centering
\caption{\xxr{Per-device performance and resource costs of the device-side confidence estimators.}}
\label{tab:device_onboarding}
\scriptsize
\renewcommand{\arraystretch}{1.15}
\resizebox{\textwidth}{!}{%
\begin{tabular}{|c|c|c|c|c|c|c|c|c|c|}
\hline
\textbf{Device}
& \textbf{Task}
& \textbf{GFLOPs}
& \textbf{Params.}
& \textbf{Infer. Latency (ms)}
& \textbf{Accuracy (\%)}
& \textbf{Offload. Rate (\%)}
& \textbf{Distill. Data}
& \textbf{Distill. Time (min)}
& \textbf{Calib. Time (s)} \\
\hline
Vivo X300 Pro
& $T_1$ (Image) & 1.82 & 12.75M & 36.10 $\pm$ 2.69
& 67.26 $\pm$ 1.01  & 49.05 $\pm$ 2.07 & 50-shot, 28 classes & 11.3 $\pm$ 0.2 & 14.5 $\pm$ 0.04 \\
\hline
Jetson Orin Nano
& $T_3$ (RGB vision) & 3.64 & 12.75M & 63.72 $\pm$ 1.91
& 43.67 $\pm$ 1.75 & 67.27 $\pm$ 2.29 & 38-shot, 15 classes & 7.2 $\pm$ 0.1 & 9.32 $\pm$ 0.26 \\
\hline
Jetson Orin NX
& $T_3$ (IR vision) & 3.64 & 12.75M & 60.89 $\pm$ 1.86
& 55.71 $\pm$ 0.64 & 79.49 $\pm$ 2.82 & 38-shot, 15 classes & 5.8 $\pm$ 0.1 & 8.86 $\pm$ 0.18 \\
\hline
Xiaomi 13 Ultra
& $T_3$ (IMU) & 0.13 & 2.19M & 7.02 $\pm$ 0.75
& 52.02 $\pm$ 0.45 & 77.49 $\pm$ 0.64 & 38-shot, 15 classes & 2.8 $\pm$ 0.1 & 4.65 $\pm$ 0.08  \\
\hline
\end{tabular}
}
\end{table*}


\begin{table}[t]
\centering
\caption{Latency and computational cost of the edge-side modules
on $T_1$ with $M_1$.}
\label{tab:module_overhead}
\scriptsize
\setlength{\tabcolsep}{4pt}
\renewcommand{\arraystretch}{1.1}
\begin{tabular}{|c|c|c|}
\hline
\textbf{Module}
& \textbf{Latency (ms)}
& \textbf{GFLOPs} \\
\hline
Modality contribution assessor
& 2.59 $\pm$ 0.05
& 0.019 \\
\hline
Feature imputation module
& 3.79 $\pm$ 0.11
& 0.003 \\
\hline
Full MFM inference
& 462 $\pm$ 4.05
& 220 \\
\hline
\end{tabular}
\end{table}

\subsubsection{\xxr{Overhead of Proposed Modules}}
\xxr{
We evaluate the device-side confidence estimators on two
smartphones and two embedded devices. As shown in
~\tabref{tab:device_onboarding}, the estimators require
0.13--3.64 GFLOPs and contain 2.19--12.75\,M parameters.
Across the four deployments, their mean on-device inference
latency ranges from 7.02 to 63.72\,ms, with mean local accuracy
of 43.67--67.26\% and mean offloading rates of
49.05--79.49\%. Supporting a new device requires re-distilling
its lightweight local estimator and recalibrating its class-wise
upload thresholds. This one-time offline adaptation uses
38--50 samples per class and takes, on average, 2.8--11.3
minutes for distillation and 4.65--14.5 seconds for threshold
calibration.
}
\xxr{
\tabref{tab:module_overhead} further reports the runtime
overhead of the edge-side modules on $T_1$ with $M_1$.
The modality contribution assessor and feature imputation
module together require only 6.38\,ms and 0.022 GFLOPs,
equivalent to 1.38\% of the full MFM latency and 0.01\% of
its computational cost, respectively. Since device-side
estimation runs concurrently with data transmission, its
latency is not fully added to the end-to-end latency. Thus,
the introduced overhead is negligible relative to the
end-to-end latency reduction achieved by \sysname.
}

\begin{figure}[t]
  \centering
  \includegraphics[width=0.36\textwidth]{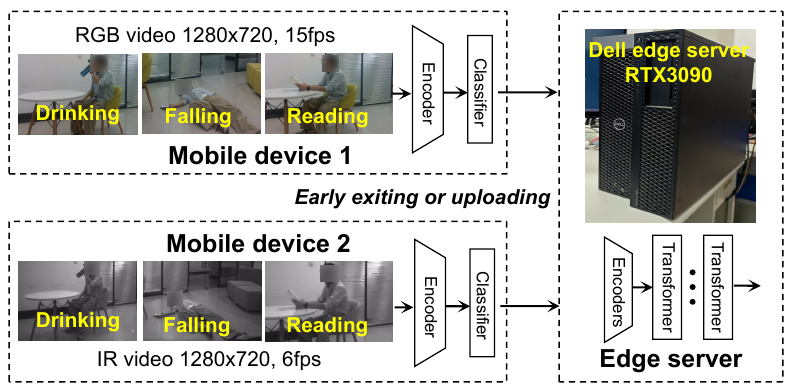}
  \vspace{-3mm}
  \caption{Illustration of the case study with mobile devices.}
  \vspace{-5mm}
  \label{fig:case_system}
\end{figure}

\begin{figure}[t]
  \centering
  \includegraphics[width=0.35\textwidth]{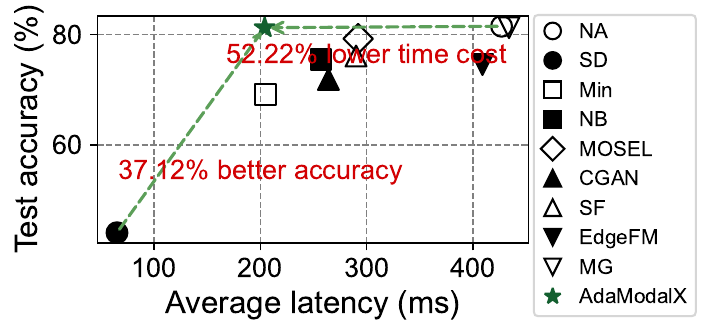}
  \vspace{-3mm}
  \caption{Comparison of accuracy \vs inference time between \sysname and other baselines in the real-world case study.}
  \label{fig:real_case_per}
  \vspace{-5mm}
\end{figure}

\subsection{Real-world Case Study}
\begin{table}[t]
\small
\centering
\caption{Comparison of FLOPs on $T_3$ with $M_1$.}
\renewcommand{\arraystretch}{1.2}
\resizebox{\columnwidth}{!}{%
\begin{tabular}{|l|c|c|c|c|c|}
\hline
\textbf{Method} & \textbf{NA} & \textbf{MG} & \textbf{CGAN} &\textbf{EdgeFM} & \textbf{\sysname} \\
\hline
GFLOPs & 335.19 & 337.01  & 335.21 & 280.24 & 235.26\\
\hline
\end{tabular}
} 
\label{tab:flops}
\end{table}

\xxr{We develop a distributed sensing system for smart-home human activity recognition covering 15 daily-life activity classes}. This system aims to enhance user experience through intelligent services, as shown in \figref{fig:case_system}.
Specifically, the system integrates a Jetson Orin Nano connected to an RGB Camera (1280x720, 15 FPS), and a Jetson Orin NX connected to an IR Camera (1280x720, 6 FPS). 
We evaluate the inference accuracy and system latency of \sysname, comparing it with nine baselines (NA~\cite{imagebind}, MG~\cite{adamml}, Min~\cite{architecting}, NB~\cite{MissingDataImputationwang2023patch}, SD~\cite{resnet}, MOSEL~\cite{mosel}, CGAN~\cite{adaflow}, SF~\cite{shared_specific}, and EdgeFM~\cite{edgefm}) under 10Mbps.
\textit{First}, as shown in \figref{fig:real_case_per}, \sysname achieves the best accuracy-latency trade-off in  distributed asynchronous multimodal sensing via importance-aware uploading and on-demand activation.
Compared with NA~\cite{imagebind}, it reduces the average latency by 52.22\% and it improves the average accuracy by 37.12\% compared with Min~\cite{architecting}.
\textit{Second}, as shown in \tabref{tab:flops}, \sysname achieves the lowest computational cost while maintaining competitive accuracy.
It reduces the computational cost by 29.81\%, 30.19\%, 29.82\%, and 16.05\% compared with NA~\cite{imagebind}, MG~\cite{adamml}, CGAN~\cite{adaflow}, and EdgeFM~\cite{edgefm}, respectively.

\begin{figure}[t]
  \centering
  \includegraphics[width=0.48\textwidth]{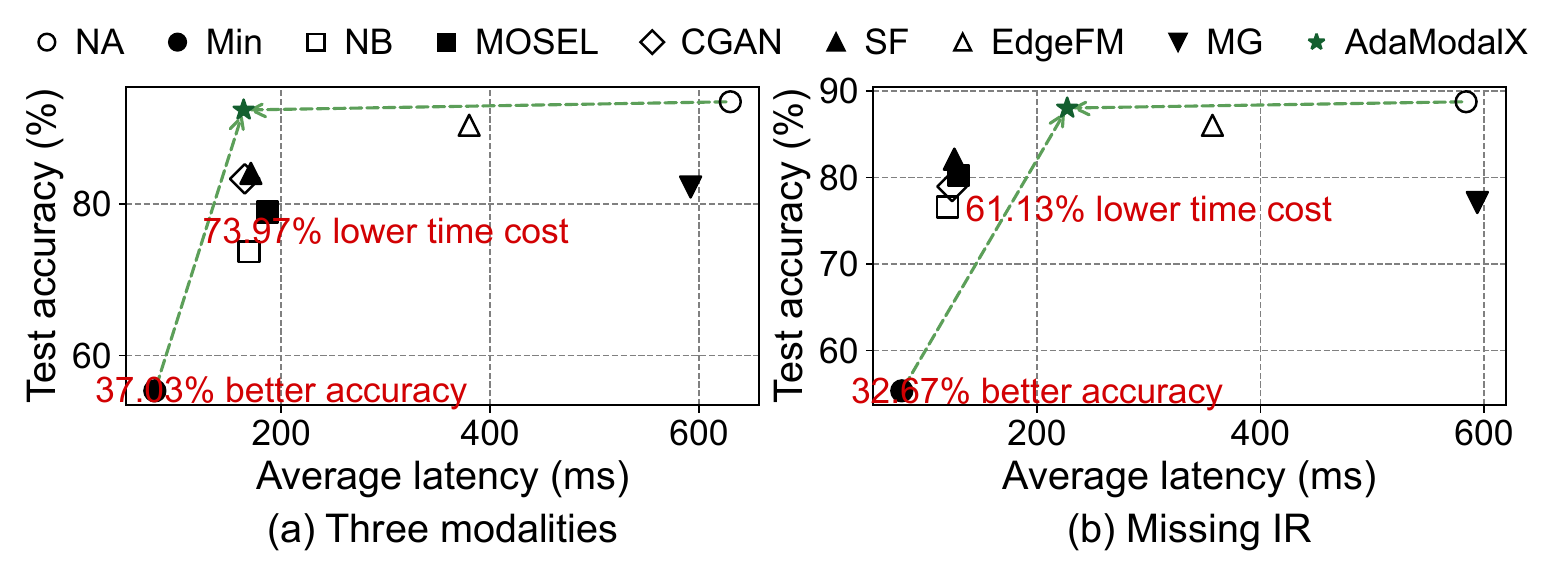}
  \vspace{-3mm}
  \caption{
  \xxr{
  Comparison of accuracy \vs latency in the three-modality human activity recognition case.
  }
  }
  \label{fig:three_modal_case}
  \vspace{-5mm}
\end{figure}

\xxr{
We further add a Xiaomi 13 Ultra to provide an IMU stream, forming a three-device RGB/IR/IMU sensing configuration under 5 Mbps.
\textit{First}, as shown in \figref{fig:three_modal_case}, \sysname achieves the best accuracy-latency trade-off under the complete three-modality setting, reducing average latency by 73.97\% over NA~\cite{imagebind} and improving accuracy by 37.03\% over Min~\cite{architecting}.
\textit{Second},  \sysname still achieves the best accuracy-latency trade-off when sensor failure makes one modality unavailable and only the remaining two modalities support inference.
}
\section{Related Work}
\label{sec:RelatedWork}
\subsection{Distributed Multimodal Sensing}
Distributed multimodal sensing systems are common in ubiquitous computing, such as autonomous driving~\cite{vips}, action recognition~\cite{RF_Camera,cat}, and intelligent healthcare~\cite{medmllm}. 
For example, VIPS~\cite{vips} proposes an infrastructure-assisted autonomous driving framework that fuses in-vehicle sensor data for improved accuracy. 
RF-Camera~\cite{RF_Camera} combines video and RFID data to enhance human gesture recognition. 
Wavoice~\cite{wavoice} integrates audio and millimeter-wave signals for precise speech recognition. 
To enhance accuracy and generalization, researchers are also exploring how MFM can be integrated into distributed sensing systems.
CAT~\cite{cat} explores how an MFM describes audiovisual events in complex environments.
Med-MFM~\cite{medmllm} enhances clinical decision-making by improving diagnostic efficiency. 
EarthGPT~\cite{earthgpt} utilizes visual enhancements and multi-granularity data for multi-sensor remote sensing.
However, they mainly focus on the development of MFMs for multimodal data or accurate inference, neglecting the inference latency caused by asynchronous data. 
In contrast, \sysname introduces \textit{importance-aware data uploading at devices} and \textit{on-demand MFM inference activation at servers} to overcome inference blocking from asynchronous data, reducing latency while maintaining accuracy.

\subsection{Asynchronous Multimodal Inference}

\xxr{
Existing approaches improve the efficiency and responsiveness of multimodal inference through \textit{modality selection}, \textit{non-blocking multimodal inference}, and \textit{collaborative inference}.
}

\xxr{
\textbf{Modality Selection.}
Modality-selection methods improve inference efficiency through \textit{policy networks} (\eg AdaMML~\cite{adamml}), \textit{modality gating} (\eg SMG~\cite{smg}), or \textit{latency-constrained selection} (\eg MOSEL~\cite{mosel}).
However, these methods make selection decisions within a request or segment without considering how modality availability changes during device--edge transmission.
They therefore do not directly address distributed mobile sensing, where modality data arrive asynchronously and become available progressively.
In contrast, \sysname progressively updates the inference candidate as modality data arrive.
}

\xxr{
\textbf{Non-blocking Multimodal Inference.}
It avoids waiting for all modalities by \textit{exploiting cross-modal features}~\cite{JointRepresentation_chen2020hgmf} or \textit{imputing missing data} (\eg AdaFlow~\cite{adaflow}, AdaFlowLite~\cite{liu2025adaflowlite}, and PATCH~\cite{MissingDataImputationwang2023patch}).
However, these methods mainly focus on using available inputs or completing unavailable modality data, without determining when to activate the edge-side MFM according to candidate utility as modalities progressively arrive.
\sysname activates edge-side MFM inference when the candidate formed by fully arrived modalities provides sufficient utility; otherwise, it imputes features from partially received inputs and re-evaluates the augmented candidate, avoiding unnecessary imputation and premature inference.
}

\xxr{
\textbf{Collaborative Inference.}
Collaborative-inference methods coordinate mobile devices and edge servers through \textit{model partitioning} ~\cite{adaknife}, \textit{data compression}~\cite{huang2022real}, or \textit{confidence-based big--small model switching} (\eg big/little~\cite{score_1}, AppealNet~\cite{appealnet}, and EdgeFM~\cite{edgefm}).
However, \textit{model partitioning} tightly couples device--edge execution and is bandwidth-sensitive, whereas \textit{data compression} incurs processing overhead but still requires waiting for delayed modalities before inference.
\textit{Big--small model switching} relies on device-side confidence estimates; inaccurate estimates may cause false early exits or unnecessary offloading.
\sysname uses joint confidence rules for device-side early exit and separately determines when to activate the edge-side MFM according to candidate utility as modality data progressively arrive.
}

\xxr{
Overall, \sysname coordinates two runtime decisions for asynchronous multimodal inference: whether to exit early based on device-side confidence and, otherwise, when to activate the edge-side MFM according to candidate utility as modality data progressively arrive.
}

\section{Conclusions}
\label{sec:conclusions}

In this paper, we address the challenge of modality asynchrony in MFM-based distributed multimodal sensing.
Unlike prior methods that require extensive retraining or architectural modifications, we introduce \sysname, a progressive multimodal inference framework that decouples inference execution from modality synchronization via runtime utility-aware orchestration. 
By learning modality utility offline and dynamically triggering inference only when sufficient data is available, \sysname achieves a balance between accuracy and latency. 
Additionally, \sysname leverages lightweight models on mobile devices to reduce data transmission and processing delays, further improving responsiveness.
Evaluations across four mobile tasks and real-world scenarios on four mobile devices demonstrate that \sysname achieves an accuracy improvement of up to 26.97\% and a system latency reduction of up to 64.75\%. 
\xxr{
Our work demonstrates the potential of \sysname in asynchronous distributed environments, although large-scale evaluations primarily emulate asynchronous modality arrivals under controlled network conditions. Future work will extend \sysname to more architectures, modalities, and devices and validate it in larger live deployments under diverse sensing and network conditions.
}

\section*{Acknowledgment}
This work was partially supported by the National Key R\&D Program of China (No.2024YFB4505502), and the National Natural Science Foundation of China (No. 62522215, 62532009, 62472354).

\bibliographystyle{IEEEtran}
\bibliography{acmart}
\vspace{-10mm}
\begin{IEEEbiography}[{\includegraphics[width=1in,height=1.25in,clip,keepaspectratio]{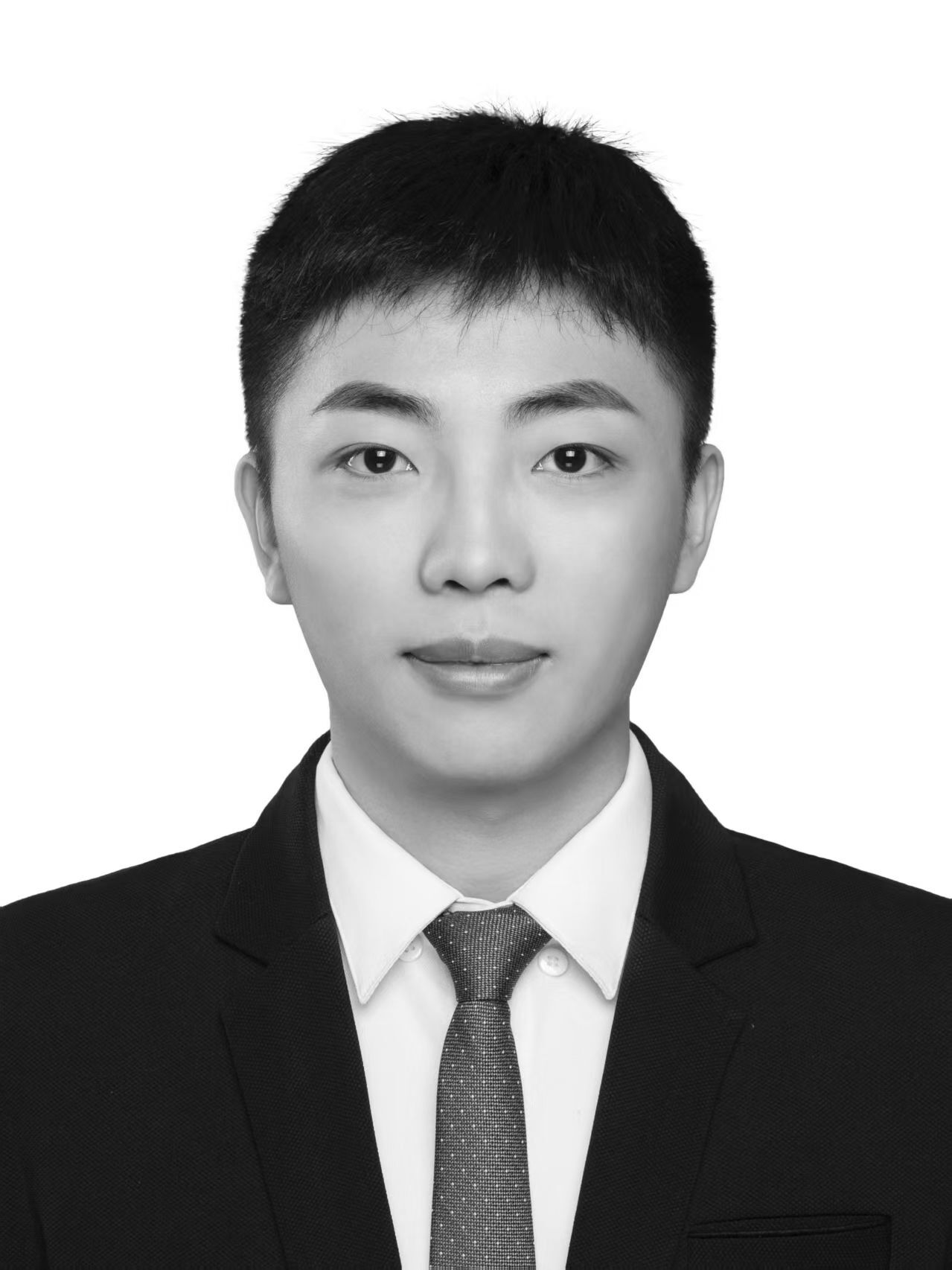}}]{Xiangrui Xu}
is a doctoral student at Northwestern Polytechnical University, China. He received B.E. from Harbin Institute of Technology in 2023. His research interests include resource-efficient mobile deep learning and mobile systems.
\end{IEEEbiography}
\vspace{-10mm}

\begin{IEEEbiography}[{\includegraphics[width=1in,height=1.25in,clip,keepaspectratio]{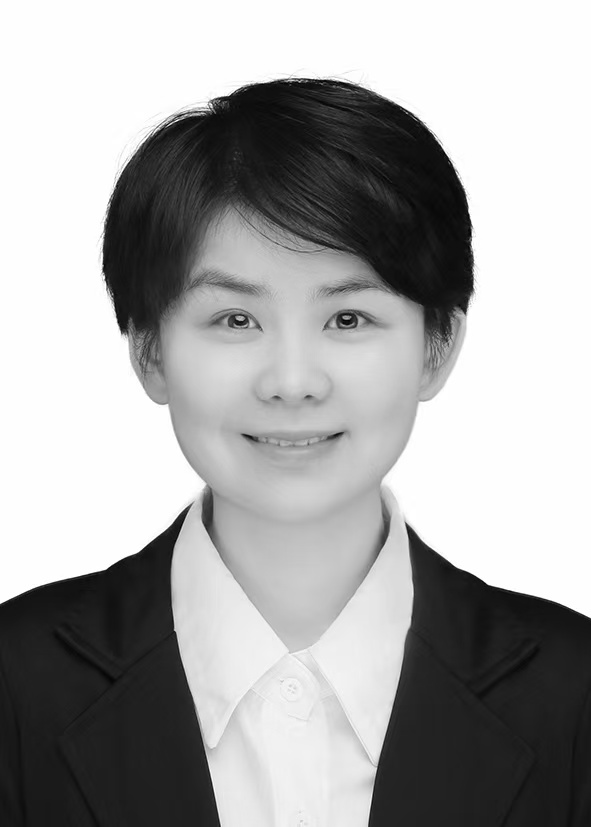}}]{Sicong Liu}
 received the PhD degree in school of computer science and technology from Xidian University in 2020. Now she is an associate professor with school of computer science, Northwestern Polytechnical University. Her research interests include mobile computing and resource-efficient mobile deep learning. She has served as the TPC member of MobiSys 2021 and BigCom 2021.
\end{IEEEbiography}
\vspace{-10mm}

\begin{IEEEbiography}[{\includegraphics[width=1in,height=1.25in,clip,keepaspectratio]{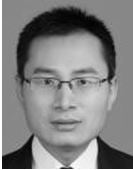}}]{Zhiwen Yu}
received the Ph.D. degree in computer science from Northwestern Polytechnical University, Xi’an, China, in 2005. He is currently the Vice President of Harbin Engineering University and a Professor at the School of Computer Science, Northwestern Polytechnical University. He was an Alexander Von Humboldt Fellow with Mannheim University, Germany, and a Research Fellow with Kyoto University, Kyoto, Japan. His research interests include ubiquitous computing, HCI, and mobile sensing and computing.
\end{IEEEbiography}
\vspace{-10mm}

\begin{IEEEbiography}[{\includegraphics[width=1in,height=1.25in,clip,keepaspectratio]{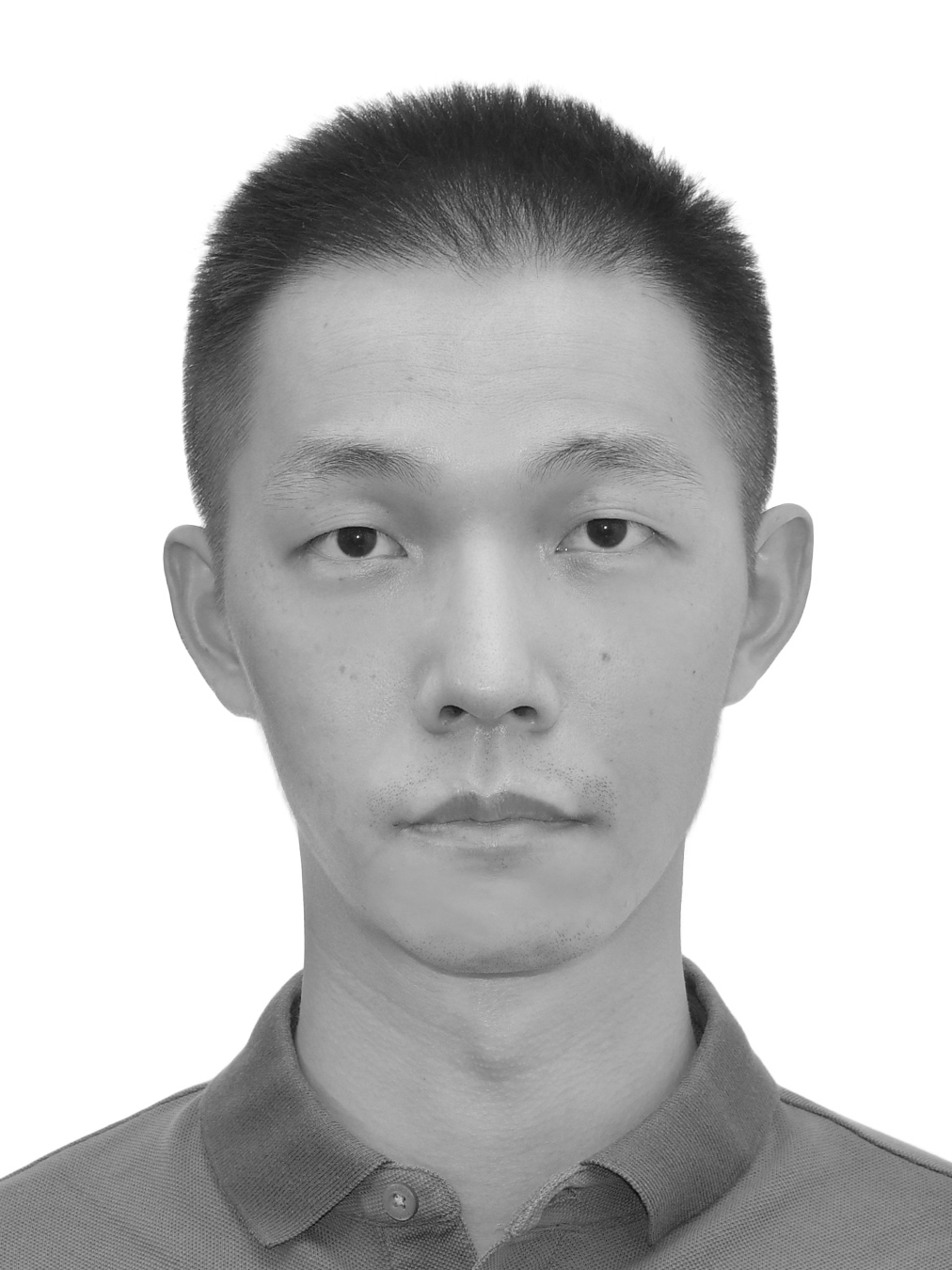}}]{Zimu Zhou}
 is currently an assistant professor in the Department of Data Science, City University of Hong Kong. He obtained his Ph.D. from the Hong Kong University of Science and Technology in 2016 and B.E. from Tsinghua University in 2011. His research interest lies broadly in mobile and ubiquitous computing, with a focus on AIoT. 
\end{IEEEbiography}
\vspace{-10mm}

\begin{IEEEbiography}[{\includegraphics[width=1in,height=1.25in,clip,keepaspectratio]{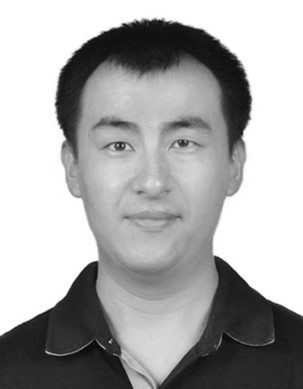}}]{Bin Guo}
 received the PhD degree in computer science from Keio University, Japan, and then was a postdoc researcher with Institut Telecom SudParis, France. He is a professor with Northwestern Polytechnical University, China. His research interests include ubiquitous computing, mobile crowd sensing, and HCI. He has served as an associate editor of the IEEE Communications Magazine and the IEEE Transactions on Human-Machine-Systems, the guest editor of the ACM Transactions on Intelligent Systems.
\end{IEEEbiography}
\end{document}